\documentclass[lettersize,journal]{IEEEtran}
\usepackage{amsmath,amsfonts}
\usepackage{algorithmic}
\usepackage{algorithm}
\usepackage{array}
\usepackage[caption=false,font=normalsize,labelfont=sf,textfont=sf]{subfig}
\usepackage{textcomp}
\usepackage{stfloats}
\usepackage{url}
\usepackage{verbatim}
\usepackage{graphicx}
\usepackage{cite}

\newcommand{\revision}[1]{{\color{black} #1}}

\usepackage[table]{xcolor}
\usepackage{booktabs}
\usepackage{multirow}
\usepackage[pagebackref,breaklinks,colorlinks,citecolor=blue,linkcolor=blue]{hyperref}

\begin{document}

\title{Improving Satellite Imagery Masking using Multi-task and Transfer Learning}

\author{
Rangel Daroya, 
Luisa Vieira Lucchese,
Travis Simmons,
Punwath Prum,
Tamlin Pavelsky,
John Gardner,\\
Colin J. Gleason, and
Subhransu Maji
\thanks{This work was supported by the NASA AIST program under Grant 80NSSC22K1487, and the National Science Foundation under Grant 2329927 and Grant 1749833. \textit{(Corresponding author: Rangel Daroya.)}}
\thanks{Rangel Daroya and Subhransu Maji are with the College of Information and Computer Sciences, University of Massachusetts, Amherst, MA 01003, USA. (email: rdaroya@umass.edu; smaji@cs.umass.edu).}
\thanks{Luisa Vieira Lucchese, John Gardner, and Punwath Prum are with the Department of Geology and Environmental Science, University of Pittsburgh, Pittsburgh, PA 15260, USA. (email: luisa.lucchese@pitt.edu; gardner.john@pitt.edu; punwath.prum@pitt.edu).}
\thanks{Travis Simmons and Colin Gleason are with the Department of Civil and Environmental Engineering, University of Massachusetts, Amherst, MA 01003, USA. (email: tsimmons@umass.edu; cjgleason@umass.edu).}
\thanks{Tamlin Pavelsky is with the Department of Earth, Marine and Environmental Science, University of North Carolina, Chapel Hill, NC 27599, USA. (email: pavelsky@unc.edu).}
}




\maketitle

\begin{abstract}
Many remote sensing applications employ masking of pixels in satellite imagery for subsequent measurements. For example, estimating water quality variables, such as Suspended Sediment Concentration (SSC) requires isolating pixels depicting water bodies unaffected by clouds, their shadows, terrain shadows, and snow and ice formation. A significant bottleneck is the reliance on a variety of data products (e.g., satellite imagery, elevation maps), and a lack of precision in individual steps affecting estimation accuracy. We propose to improve both the accuracy and computational efficiency of masking by developing a system that predicts all required masks from Harmonized Landsat and Sentinel (HLS) imagery. Our model employs multi-tasking to share computation and enable higher accuracy across tasks. We experiment with recent advances in deep network architectures and show that masking models can benefit from these, especially when combined with pre-training on large satellite imagery datasets. We present a collection of models offering different speed/accuracy trade-offs for masking. MobileNet variants are the fastest, and perform competitively with larger architectures. Transformer-based architectures are the slowest, but benefit the most from pre-training on large satellite imagery datasets. Our models provide a 9\% F1 score improvement compared to previous work on water pixel identification. When integrated with an SSC estimation system, our models result in a 30x speedup while reducing estimation error by 2.64 mg/L, allowing for global-scale analysis. We also evaluate our model on a recently proposed cloud and cloud shadow estimation benchmark, where we outperform the current state-of-the-art model by at least 6\% in F1 score.

\end{abstract}

\begin{IEEEkeywords}
transfer learning, multi-task learning, global surface water detection, deep learning, suspended sediment
\end{IEEEkeywords}

\section{Introduction}
\label{sec:intro}

\IEEEPARstart{I}{solating} different types of pixels is a pre-requisite in many remote sensing tasks. Prior to estimating a variable of interest, it is often crucial to isolate pixels covering specific land cover types such as water~\cite{li2003remote, tilton2006utilizing}, forests~\cite{huang2010automated, aravena2023high}, agriculture~\cite{atzberger2013advances, valero2016production}, urban areas~\cite{huang2018dmblc, xiao2004using}), or remove artifacts such as clouds and shadows~\cite{zhang2024improvedlana, huang2010automated}. Landsat and Sentinel-2 data come with water masks processed with the function of mask (Fmask) algorithm~\cite{qiu2019fmask, zhu2015improvement}, but these are often insufficient due to the tendency of Fmask to miss clouds~\cite{zhang2024improvedlana}, misclassify water in complex environments~\cite{chen2020novel}, and misclassify clear ice as open water~\cite{yang2021simple}. Therefore, remote sensing workflows across many domains have developed specific preprocessing and masking tasks essential in estimating a quantity of interest~\cite{gardner2021color, zhang2024improvedlana, dethier2020toward}.

In this paper, we demonstrate a new approach to masking for a case of estimating suspended sediment concentration (SSC). Although SSC is discussed here as an application, our proposed masking procedure is broadly applicable to other remote sensing estimation methods. SSC can be estimated in rivers from a given satellite image by first classifying the pixels into water and non-water classes to identify valid areas for analysis~\cite{li2003remote, gardner2023human, dethier2020toward}. All artifacts from clouds, cloud shadows, terrain shadows, snow, and ice, considered as non-water, are also typically removed so that only ``good quality" water pixels remain~\cite{dethier2020toward}.  Once good quality pixels are identified, summary statistics such as mean, variance, and counts from the reflective bands of water pixels are used as input to train a model to predict SSC~\cite{zhang2014suspended, langhorst2022simultaneous, narayanan2024riverine}. Threshold-based classification of pixels such as via the Fmask algorithm \cite{qiu2019fmask} and the modified normalized difference water index (MNDWI) \cite{xu2006modification} can be effective for water identification, but they are highly dependent on the locations and weather conditions where thresholds are determined for the different types of masks~\cite{du2016water}. Fmask applies spectral tests to thermal and optical bands for identifying cloud, shadow, water, and snow pixels with thresholds based on the global optima across sampled reference images~\cite{qiu2019fmask}. However, a lower threshold for identifying clouds that works well in areas with predominantly thin clouds could result in overestimation of clouds in areas with no clouds at all. Similar issues arise with other masks, where the distribution of dark surfaces (e.g., burned areas, wetlands) could affect the prediction of shadows and water due to similar spectral characteristics. MNDWI thresholds the ratio involving the green and short-wavelength infrared (SWIR) bands since radiation from SWIR is strongly absorbed by water~\cite{xu2006modification}. However, it is also sensitive to shadows from different topographic conditions~\cite{donchyts201630} and is very sensitive to snow; it is identical in form to the normalized difference snow index~\cite{salomonson2004estimating}.

Once water is obtained (Fmask also returns snow, ice, and clouds following the same threshold-based logic), water pixels must be sorted into those of `good' quality. The definition of `good' is subjective, but often sun glint and shadows are problematic for reflectance based applications~\cite{greenberg2022improved, gao2021correction}. Ancillary data are needed to model terrain shadows. Shadows can be obtained using the approximate position of the sun (based on the date and time of day) and the surrounding topography (using digital elevation maps (DEM)). DEMs are large files, especially those that are global~\cite{yamazaki2017high}, so this shadowing process is data intensive. While we have given examples of preprocessing for SSC estimation, the same issues apply for other applications that require the isolation of pixels to use the surface reflectance to drive another estimation process.

Deep learning networks provide an alternative to threshold-based methods; previous work has shown the promising performance of deep learning models for segmenting images where each pixel is classified to a specific class \cite{zhang2023delivering, xu2023side, isikdogan2017surface, chai2019cloud}. \revision{In particular, semantic segmentation has been explored together with deep learning methods similar to~\cite{chai2019cloud} where a fully convolutional neural network is used to detect cloud and cloud shadows. } DeepWaterMap (DWM) \cite{isikdogan2017surface} uses a convolutional neural network (CNN) to segment water pixels from Landsat data. LANA~\cite{zhang2024improvedlana} uses an improved CNN with attention-based mechanisms~\cite{vaswani2017attention} to identify cloud and shadow pixels from Landsat 8 data. Other architectures such as DeepLabv3+~\cite{chen2018encoder}, MobileNet~\cite{howard2019searching}, SegNet~\cite{badrinarayanan2017segnet}, Vision Transformers~\cite{dosovitskiy2020image}, and Swin Transformers~\cite{liu2021swin} have shown encouraging results for object-based pixel classification in images. 

Apart from architecture considerations, deep learning methods require a sufficient amount of labeled training data to achieve competitive performance~\cite{sun2017revisiting, schuhmann2022laion, deng2009imagenet}. To reduce dependency on large datasets, transfer learning is applied where models are first trained on a different task (e.g., image classification) with a large dataset (e.g., ImageNet~\cite{deng2009imagenet} with more than 14M images) and then fine-tuned on a specific task (e.g., masking)~\cite{girshick2015fast, long2015fully, he2017mask}. Several large datasets such as ImageNet~\cite{deng2009imagenet}, Prithvi~\cite{Prithvi-100M-preprint}, and Satlas~\cite{bastani2023satlaspretrain} have become available with millions of labeled training images that were shown to be effective for pre-training. At the same time, fine-tuning can be applied using the recently released Dynamic Surface Water eXtent (DSWx)~\cite{jones2019improved} product associated with the Harmonized Landsat-Sentinel (HLS) project~\cite{claverie2018harmonized} contains labeled masking data with high frequency and near-global coverage. It has been used for flood detection, wildfire mapping, and reservoir monitoring using changes in water bodies and vegetation, and can potentially be used for masking applications (e.g, water, cloud, shadow identification)~\cite{bato2023first}. While DSWx can be used as the sole source of masks, it is limited by the required input sources to produce them (Copernicus DEM, Copernicus land cover, ESA worldcover, NOAA GSHHS shapefile, and HLS). DSWx is currently not available for historical data and it would be challenging and resource-intensive to generate~\cite{jones2019improved}. In contrast, deep learning models can predict masks without relying on multiple sources (e.g., predict masks using only HLS).

With the increasing availability of datasets for both pre-training and task-specific fine-tuning, deep learning models show great potential to find patterns and generalize on unseen data for satellite imagery masking. However, these architectures typically use a single model to predict a single output. In the case of predicting five masks (i.e., water, cloud, cloud shadow, snow/ice, terrain shadow) as is required for SSC estimation, five such models would be necessary, which would require more time and resources for training and inference. 
Multi-task models~\cite{caruana1997multitask} introduce a framework that allows the simultaneous prediction of multiple outputs (e.g., water, shadow, cloud masks). It was shown to improve generalization performance by learning multiple tasks at the same time, while simplifying the training to a single model~\cite{zhang2018learning}.  Instead of training five separate models where each model predicts a single type of mask (e.g., water), a multi-task framework uses a single model to predict all five masks (water, cloud, cloud shadow, terrain shadow, snow/ice). Thus, approximately five times less resources would be needed for training and inference.

In this paper, we propose an end-to-end framework for estimating suspended sediment concentration using a multi-task deep learning model. We investigate the reliability and efficiency of identifying water pixels and other artifacts (e.g., clouds, cloud shadows, terrain shadows, snow/ice) from satellite imagery using multi-task models and \revision{transfer learning through }different pre-training datasets (e.g., ImageNet, Satlas, Prithvi), and compare our results with existing models such as LANA~\cite{zhang2024improvedlana} and DWM~\cite{isikdogan2019seeing}. We also compare performance of combining single-task models against the multi-task equivalent. Finally, we show the impact of multi-tasking by evaluating the effect of masking improvements on downstream tasks -- here using SSC estimation as a case study.

\section{Data Sources}
\label{sec:data}

\subsection{Harmonized Landsat-Sentinel}

The Harmonized Landsat-Sentinel (HLS) project \cite{claverie2018harmonized} uses the Operational Land Imager (OLI) and Multi-Spectral Instrument (MSI) sensors from the Landsat and Sentinel remote sensing satellites. The combined temporal frequency is 2-3 days at 30m spatial resolution with Landsat data starting from February 2013, and Sentinel-2A/2B starting from June 2015/March 2017. The satellites cover all land areas in the globe except Antarctica. The HLS data provides 15 harmonized bands as enumerated in Table~\ref{table:hls-bands}.

\begin{table}[!t]
    \small
    \begin{center}
    \setlength{\tabcolsep}{2pt}
    \caption{The 15 harmonized bands present in HLS
    }
    \label{table:hls-bands}
    \begin{tabular}{l | c}
    \toprule
    Band & Wavelength\\
    \midrule
    Coastal Aerosol & $0.32-0.45 \mu m$\\
    Blue & $0.45-0.51 \mu m$ \\
    Green  & $0.53-0.59 \mu m$ \\
    Red  & $0.64-0.67 \mu m$ \\
    Red-edge 1  & $0.69-0.71 \mu m$ \\
    Red-edge 2  & $0.73-0.75 \mu m$ \\
    Red-edge 3  & $0.77-0.79 \mu m$ \\
    Near Infrared (NIR) broad  & $0.78-0.88 \mu m$ \\
    NIR narrow  & $0.85-0.88 \mu m$ \\
    Short-Wave Infrared 1 (SWIR-1) & $1.57-1.65 \mu m$ \\
    SWIR-2 & $2.11-2.29 \mu m$ \\
    Water Vapor & $0.93-0.95 \mu m$ \\
    Cirrus & $1.36-1.38 \mu m$ \\
    Thermal Infrared 1 & $10.60-11.19 \mu m$ \\
    Thermal Infrared 2  & $11.50-12.51 \mu m$ \\
    \bottomrule
    \end{tabular}
    \end{center}
\end{table}

Input features for training the model were derived from HLS -- the bands Blue, Green, Red, Near Infrared (NIR), SWIR1, SWIR2 were used, similar to DeepWaterMap \cite{isikdogan2017surface}.

\subsection{Annotated Global Surface Water Masks}
\label{subsec:data-water-masks}
More than 3 million tiles are available in the Dynamic Surface Water eXtent (DSWx) products through NASA's Earth Observing System Data and Information System (EOSDIS)~\cite{jones2019improved}. DSWx products cover inputs generated from Sentinel-1, NISAR, and HLS. This work focuses primarily on the DSWx product that uses HLS as the image-based input. The product provides a map of the extent of surface waters across all landmasses excluding Antarctica. Each tile has ten layers, where each layer has a size of 3660 $\times$ 3660 pixels (each pixel has a 30m resolution).

Labels were obtained from a year's worth of DSWx data from April 2023 to March 2024 to train a model to predict the different masks. Labels included the five masks for water, cloud, cloud shadow, snow/ice, and terrain shadow (from DEM~\cite{yamazaki2017high}). Only high confidence water pixels (both partial and open water) were used as water mask labels based on the DSWx product. The time frame was selected to obtain samples from varying seasons and weather conditions. Each label from the DSWx tiles were matched to available HLS tiles, and sampled as follows:

\begin{figure*}[!t]
    \begin{center}
    \includegraphics[scale=0.7]{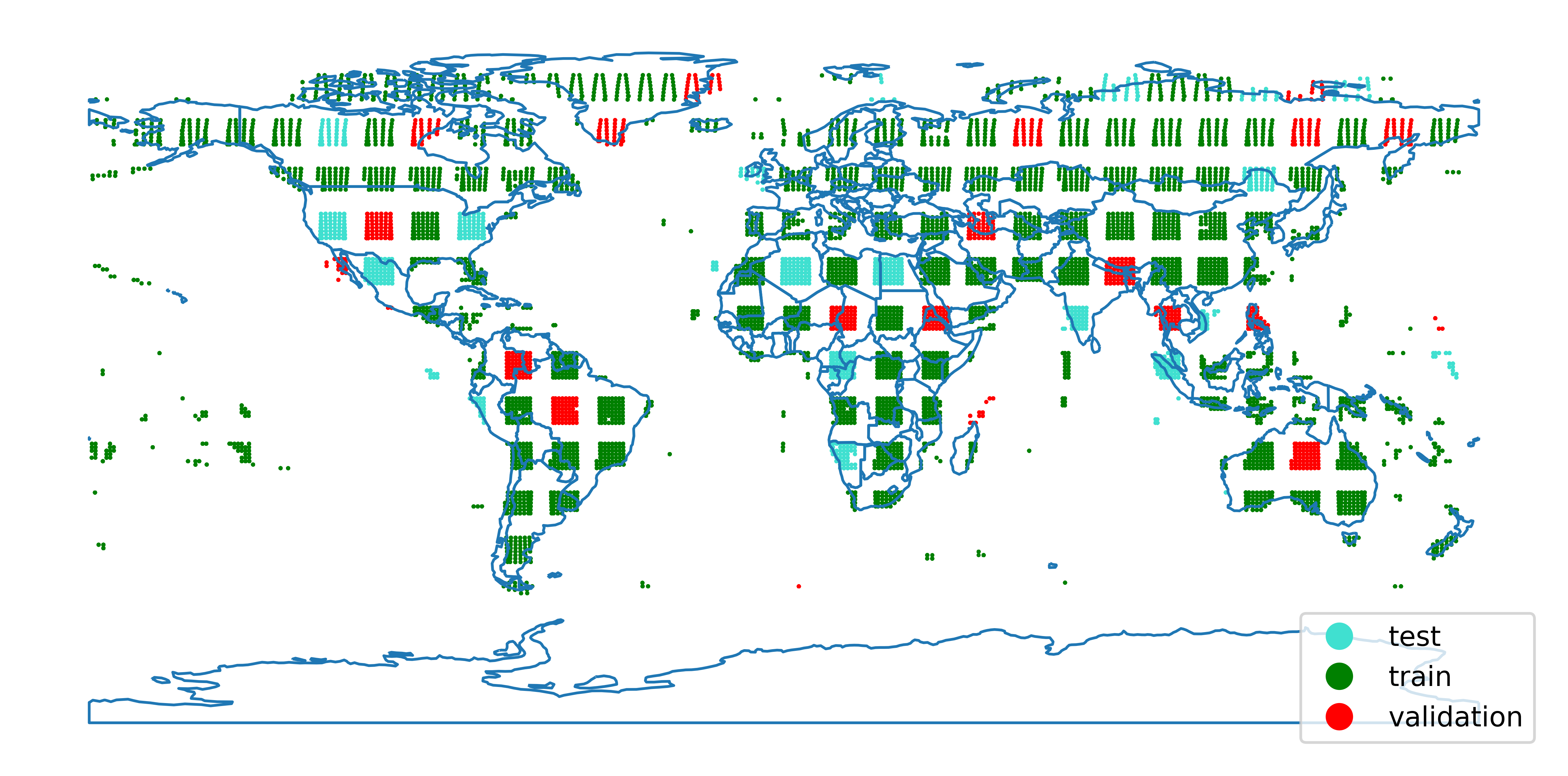}
    \end{center}
    \vspace{-6mm}
    \caption{\textbf{Geographic distribution of train, validation, and test data.} The dataset has a global coverage with train, validation, and test splits spanning different locations. Each dot represents the center of the sampled tile. Gaps are present to prevent overlap between different data splits, since each tile has a coverage of 109.8 km × 109.8 km. Following the computer science convention, training set is used for updating model weights during training, validation set is used for selecting hyperparameters, and test set is not seen by the model except when evaluating performance.
    }
    \label{fig:data-distrib}
\end{figure*}

\begin{table*}[!t]
    \small
    \begin{center}
    \caption{Average number of pixels per image per type of mask across all samples in the data splits
    }
    \label{table:data-ave-pixels}
    \begin{tabular}{l | c c c c c}
    \toprule
    Data Split & Water & Cloud Shadow & Cloud & Snow/Ice & Terrain Shadow \\
    \midrule
    Train & 51,042 & 43,709 & 229,906 & 32,318 & 545,703 \\
    Validation & 42,142 & 40,588 & 210,917 & 39,278 & 547,651 \\
    Test & 36,213 & 40,916 & 213,802 & 22,397 & 552,310 \\
    \bottomrule
    \end{tabular}
    \end{center}
    \vspace{-6mm}
\end{table*}
\begin{itemize}
    \item Temporally: for each tile, 20 data points were uniformly sampled throughout the year to cover different seasons and weather conditions. This is roughly 1 sample every 2 weeks.
    \item Spatially: The globe was subdivided onto a $6^{\circ} \times 6^{\circ}$ grid. All available tiles contained in each cell were used. This process was done to make sure that adjacent cells would have no overlapping tiles.
\end{itemize}

In addition, satellite tiles that did not contain data (i.e., all pixels flagged as having no data) were removed from the dataset. Samples with incomplete labels were also removed such that samples that had all 5 masks remained. Each HLS feature paired with a DSWx label was cropped to a size of 512 $\times$ 512 pixels. We applied spatial validation on the data to prevent data leakage and to measure the predictive performance of the model on unseen locations. Train, validation, and test data points were based on the grid defined during sampling. Fig.~\ref{fig:data-distrib} shows the distribution of the sampled data. The extraction and processing resulted in 107,250 labeled data -- 82,247 for training, 12,849 for validation, and 12,154 for testing. Throughout the paper, we use the computer science convention for defining train, validation, and test splits. The training samples are used for training the model, the validation samples are used for selecting hyperparameters and thresholds (if any), and the testing samples are exclusively used to evaluate the performance of the models. Table~\ref{table:data-ave-pixels} shows the average number of pixels per image sample across the different data splits. Different classes across training, validation, and test have a similar number of pixels per mask.

\subsection{In-situ Suspended Sediment Concentrations}
Suspended sediment concentrations (SSC) were obtained from water quality databases~\cite{dethier2019hydroshare, dethier2020toward} across different locations around the globe. Data were collected in-situ in different sites in the United States (US Geological Survey, 2018), Canada (The Water Survey of Canada, 2018), South America (Agência Nacional de Águas, 2017), Taiwan (Taiwan Water Resource Agency, 2018), and Europe (European Environment Agency, 2020). Additional databases such as GEMStat~\cite{farber2018water} and Glorich~\cite{hartmann2014brief} also have metadata that indicate the quality of the measurements and the depth at which the concentrations were measured. 

\revision{The SSC data were obtained from the United States, Canada, South America, Europe, and parts of Asia (e.g., Taiwan, Japan, China). A total of 244,000 in-situ SSC data were obtained. HLS tiles and in-situ SSC values were then matched based on the location and date. Satellite images taken within one day of the sampled SSC value, and images with nonzero data were used. Using this matching criteria, we obtained 24,328 data points with both in-situ SSC values and corresponding HLS tiles. These data points were split into train, validation, and test with 50\%, 25\%, and 25\% of the data, respectively. These three sets were sampled to be in different spatial locations to prevent data leakage, and to ensure an accurate assessment of the model’s generalizability. The final set of SSC values range from 0.003 to 723.0 mg/L, taken from April 2013 to October 2021~\cite{lucchese2024modeling}.} 

\revision{We removed samples that could not be reliably measured. The metadata of each SSC sample includes the minimum measurable amount at that location, which is defined by the specific sensor used for SSC measurement. We define reliable samples as those that are above the given minimum measurable amount. We additionally only use samples near the river surface taken at a depth of at most one meter, since the penetration of light---used for obtaining image data---is at most only a few meters (less for turbid water).}

\section{Methods}
\label{sec:methods}

\subsection{Multi-task Model}

\begin{figure}[!t]
    \begin{center}
    \includegraphics[scale=0.45]{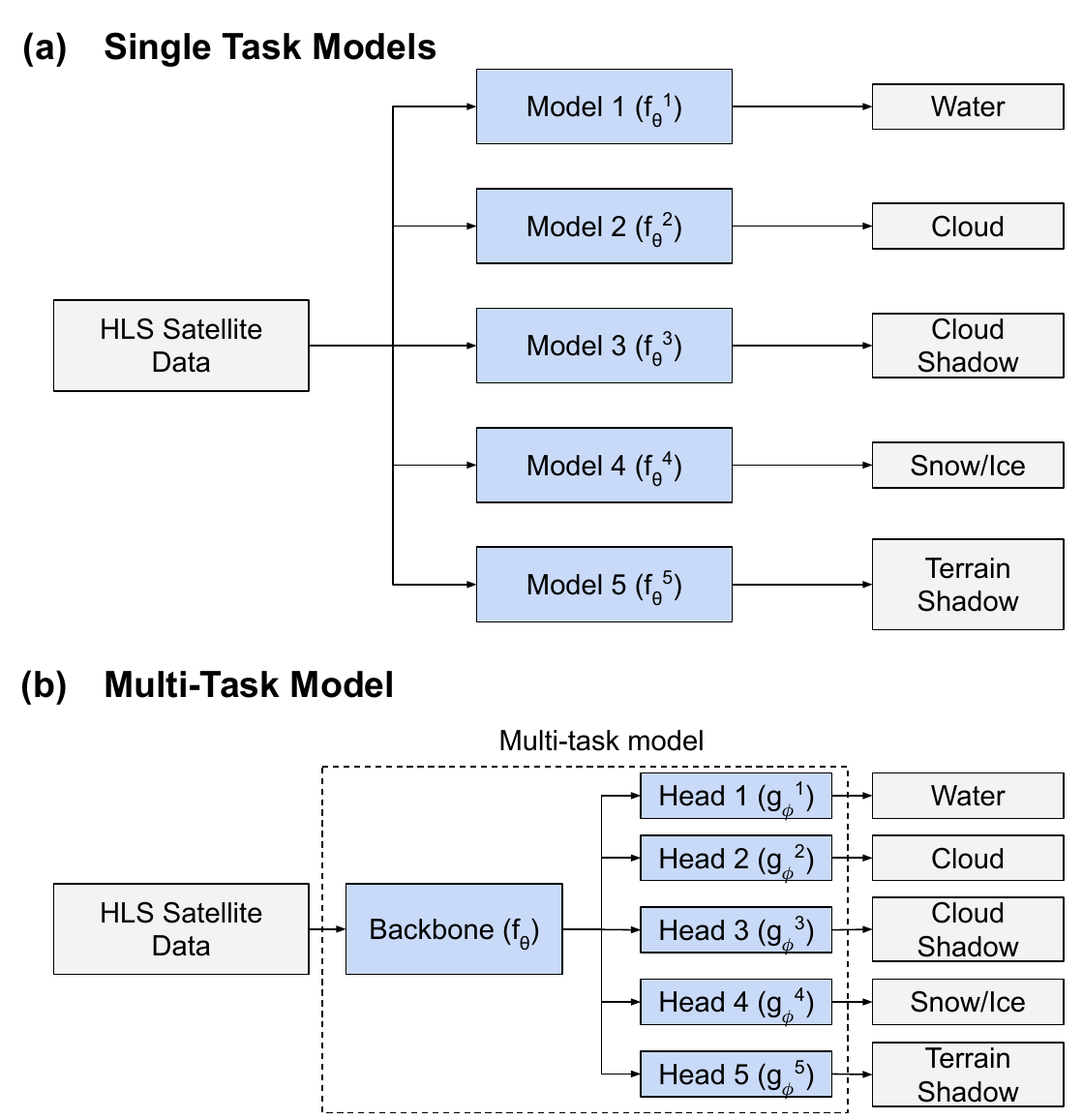}
    \end{center}
    \caption{\textbf{Comparison of multiple single-task models and multi-task model.} (a) Evaluating a model for each output is resource intensive, since it would require running five separate models. (b) Shows a multi-task model setup where only one model is predicting all five outputs at the same time, using approximately one-fifths of the resources for training.
    }
    \label{fig:single-vs-multitask}
\end{figure}

Fig.~\ref{fig:single-vs-multitask} shows how single task models differ from multi-task models. When predicting five masks, single task models would require five separate models, each trained and run separately. At the same time, five models are stored, requiring approximately five times as much storage compared to having a single model that predicts all masks at the same time. The multi-task model trains on five different masks with one single backbone model that extracts a general feature. From a general feature, five small ``heads" for each of the masks is trained that is composed of a few trainable layers (much smaller than the backbone). Instead of five large models trained separately, the multi-task model only has to train one large model and five small sets of trainable layers simultaneously. The multi-task framework would thus save time and resources for training and inference.

More formally, a single model is a parameterized function $f$ that transforms an input $x$ into output $\hat{y}$ by learning the function parameters $\theta: f_{\theta}(x) = \hat{y}$. The parameters $\theta$ are learned by training $f$ on a data set of $n$ image-label pairs using a specified loss function. Given a training data set $\mathcal{D}=\{(x_i,y_i)\}_{i=1}^n$ where $x_i$ is the input image and $y_i$ is the corresponding label, the goal is to have the model output $f_{\theta}(x_i)=\hat{y}_i$ be as close as possible to $y_i$. In our work, $x_i \in \mathbb{R}^{512 \times 512 \times 6}$ is the $512 \times 512$ pixel image composed of six bands from HLS (red, green, blue, NIR, SWIR-1, SWIR-2), and $y_i \in \mathbb{R}^{512 \times 512 \times 1}$ is one of water, cloud, cloud shadow, snow/ice, or terrain shadow mask. For a multi-task model that outputs five masks at the same time, there would be five labels $y_i^{m}$ and five corresponding model outputs $\hat{y}_i^{m}$ where $m \in [$water, cloud, cloud shadow, snow/ice, terrain shadow$]$. The multi-task model can then be represented as follows:
\begin{equation}
    z_i = f_{\theta}(x_i)
    \label{eq:feat-multitask}
\end{equation}
\begin{equation}
    y_i^{m} = g_{\phi_m}^{m}(z_i)
    \label{eq:output-multitask}
\end{equation}

Equation~\ref{eq:feat-multitask} computes the general feature used for all masks in the model. To compute each mask $m$, there are five smaller networks $g^m$ parameterized by $\phi_m$ -- the networks have the same structure but have different parameters for each mask $m$ -- so that each output mask can be obtained through Equation~\ref{eq:output-multitask}. Both $f$ and $g$ are trained simultaneously using the loss function in Equation~\ref{eq:loss} where $\mathcal{L}_{bce}$ is the binary cross entropy loss computed for each of the five labels. $\mathcal{L}$ is minimized by optimizing the parameters $\theta$ and $\phi_m$ for all $m$.

\begin{equation}
    \mathcal{L} = \frac{1}{N}  \sum_{i=1}^N \sum_{m} \mathcal{L}_{bce} \left(\hat{y}_i^m, y_i^m \right)
    \label{eq:loss}
\end{equation}

The binary cross entropy loss $\mathcal{L}_{bce} \in [0,\infty)$ can be computed by comparing all pixels in the prediction $\hat{y}_i^m$ and the ground truth $y_i^m$. For an image with $W \times H$ pixels, Equation~\ref{eq:bce} shows the loss between the predicted and ground truth for mask $m$. The variable $y_{i,(j,k)}^m \in \{0,1\}$ is the label or ground truth value for the pixel in position $(j,k)$, and $\hat{y}_{i,(j,k)}^m \in [0,1]$ is the predicted probability of the mask for the pixel in position $(j,k)$.
\begin{multline}
    \mathcal{L}_{bce}\left(\hat{y}_i^m, y_i^m \right) = \frac{1}{WH} \sum_{j=1}^W \sum_{k=1}^H - \Biggl(y_{i,(j,k)}^m \log \hat{y}_{i,(j,k)}^m + \\ \left(1-y_{i,(j,k)}^m \right) \log\left(1-\hat{y}_{i,(j,k)}^m \right) \Biggr)
    \label{eq:bce}
\end{multline}

Using a multi-task model, we explore different architectures and setups for the backbone which is crucial for extracting useful features to predict the different masks simultaneously.

\subsubsection{Network Architectures}
Convolutional Neural Networks (CNNs)~\cite{lecun1998gradient} and Transformers~\cite{dosovitskiy2020image} are two main types of model architectures that have recently emerged for various computer vision tasks such as masking/segmentation~\cite{baheti2020semantic, badrinarayanan2017segnet}, image classification~\cite{guo2017simple, dosovitskiy2020image}, and object detection~\cite{girshick2015fast, zhu2020deformable}. While both architectures are effective, they have different compositions. CNNs use a learnable kernel that slides across an image to extract features~\cite{lecun1998gradient}; transformers~\cite{liu2021swin, dosovitskiy2020image} decompose an image into patches and use attention \cite{vaswani2017attention} to learn global dependencies and better context across the whole image. We describe both architectures in more detail below.

\noindent \textbf{Convolutional Neural Networks.} CNNs are networks where the model $f$ encodes translational invariance and spatial locality. Translational invariance and spatial locality are achieved through the convolution layer, a key operator in CNNs. The convolution layer learns a kernel $K \in \mathbb{R}^{k \times k}$ applied to an image $I \in \mathbb{R}^{W \times H}$. Equation~\ref{eq:convolution} shows the operation where $S_{ij}$ is the output of the layer at the input image pixel location $(i,j)$. The same operation is applied to all pixel positions $(i,j)$ in $I$ to form a feature map $S \in \mathbb{R}^{(W-k+1) \times (H-k+1)}$. Other layer parameters such as stride, dilation, and padding can also control the size of the feature map.

\begin{equation}
    S_{ij} = (I \ast K)_{ij} = \sum_{a=0}^{k-1} \sum_{b=0}^{k-1} I_{i+a,j+b} K_{a,b} \quad \forall i,j
    \label{eq:convolution}
\end{equation}

In addition to convolution layers, other operators such as pooling layers, nonlinear activations, and fully connected layers are also used to create CNNs. Pooling layers contribute to the translational invariance of CNNs by dividing the input image into patches, and aggregating each patch (e.g., by taking the maximum) to produce smaller feature maps. Nonlinear activations apply nonlinear transformations to the input to learn complex relations between inputs and outputs. Some examples include rectified linear unit: $\text{ReLU}(x)=\text{max}(0,x)$ \cite{nair2010rectified} and sigmoid: $\sigma (x) = 1/(1 + \exp (-x))$ \cite{maas2013rectifier}. Fully connected layers apply a linear transformation to all pixels in the input to produce a value in the output. Multiple linear transformations could be applied simultaneously to all input pixels to output multiple values.

CNNs are created by stacking a series of layers described above. Typical sequences are composed of multiple blocks of convolution - pooling - nonlinear. DeepLabv3+~\cite{chen2018encoder}, MobileNet~\cite{howard2019searching}, SegNet~\cite{badrinarayanan2017segnet}, U-Net~\cite{ronneberger2015u}, and ResNet~\cite{he2016deep} are commonly used architectures for various vision applications (e.g., object detection, segmentation, classification) that use a combination of these layers.

\noindent \textbf{Transformers.} Transformers do not have inductive biases on spatial locality that are present in CNNs due to the latter's use of sliding window kernels. As a result, transformers can learn global features that could perform better than CNNs \cite{dosovitskiy2020image}. The attention layer \cite{bahdanau2014neural} is a critical block in transformers that models long-range dependencies in images to learn a global representation. To apply it, the input image is divided into $n$ patches and flattened into a $d$-dimensional vector to have an input $X \in \mathbb{R}^{n \times d}$. Each attention layer learns a set of query ($W^Q \in \mathbb{R}^{d \times d_q}$), key ($W^K \in \mathbb{R}^{d \times d_k}$), and value ($W^V \in \mathbb{R}^{d \times d_v}$) weight matrices. These are applied to the input to produce queries $Q=XW^Q$, keys $K=XW^K$, and values $V=XQ^V$. The output of the attention layers is then computed (Equation~\ref{eq:attention}). Additional learnable parameters such as positional embeddings are also added to the input patches to give information on the original position of the image patch~\cite{dosovitskiy2020image}.

\begin{equation}
    \text{Attention}(Q,K,V) = \text{softmax} \left( \frac{Q K^T}{\sqrt{d_k}}\right) V
    \label{eq:attention}
\end{equation}

Similar to CNNs, a transformer model $f$ stacks a series of blocks and use other layers such as nonlinear activations and fully connected layers. Vision Transformers (ViT) \cite{dosovitskiy2020image} and Swin Transformers (Swin-T) \cite{liu2021swin} are commonly used architectures applied to vision tasks such as classification and segmentation.

\subsubsection{Transfer Learning}
Deep learning models have millions of parameters that need to be optimized during training. When only a small dataset is available for training, there is a risk of model overfitting, where the model simply memorizes the training data and does not learn to generalize to unseen samples. To mitigate this risk, transfer learning first trains the model on a very large dataset and then fine-tunes it on a separate dataset for the specific application~\cite{girshick2015fast, long2015fully, he2017mask}. Training on the larger dataset teaches the model to extract useful feature maps from images that could be helpful for a general understanding of images, which could then be used for applications such as classification~\cite{girshick2015fast}.

Building on this idea, \revision {we apply transfer learning by first pre-training} the backbone model $f_{\theta}$ (Fig.~\ref{fig:single-vs-multitask}b) on a different but much larger dataset to learn the parameters $\hat{\theta}$. Then, we replace the last layer with the task-specific heads ($g_{\phi}$) and train the model in Fig.~\ref{fig:single-vs-multitask}b \revision {using the annotated global surface masks (introduced in \S~\ref{subsec:data-water-masks}) and} with the backbone parameters $\theta$ initially set to $\hat{\theta}$. Unlike other methods that start the optimization from random parameters, we optimize the multi-task model starting from parameters that can already extract acceptable features for understanding images \revision{(i.e., we apply transfer learning from other larger existing datasets)}.

ImageNet~\cite{deng2009imagenet}, Satlas~\cite{bastani2023satlaspretrain}, and Prithvi \cite{Prithvi-100M-preprint} were explored for pre-training in this work. ImageNet is an image classification dataset of 1,000 object classes containing more than 1 million training images, 50,000 validation images, and 100,000 test images. Pre-training on ImageNet has shown impressive results on image classification \cite{donahue2014decaf}, image segmentation \cite{dai2016instance}, and object detection \cite{girshick2014rich}. Satlas was recently introduced as a remote sensing dataset that combines images from Sentinel-2 and National Agriculture Imagery Program (NAIP) to produce 302 million labeled images for various tasks including classification, regression, object detection, and segmentation. Their work shows competitive performance against other ImageNet-pretrained models on remote sensing related tasks. Prithvi was pre-trained on HLS data for the contiguous US using a transformer architecture. It was shown to perform well on flood mapping and burn scar segmentation. In our work, ImageNet, Satlas, and Prithvi are explored as pre-training methods \revision {as a way to apply transfer learning, and} as a starting point before further training on the annotated DSWx dataset introduced in \S~\ref{subsec:data-water-masks}.

\subsection{Baselines for Masking}
We describe the methods previously used to find water and cloud pixels from satellite data to compare the performance of our proposed model. Fmask and MNDWI are based on finding empirically derived thresholds to isolate cloud and water pixels. Deep learning methods~\cite{isikdogan2017surface,wieland2019multi,zhang2024improvedlana} train a model to predict pixels similar to our proposed model. We discuss each of these below.

\subsubsection{Fmask}
The Fmask algorithm \cite{qiu2019fmask} uses the reflective bands and brightness temperatures from HLS to compute the probability of each pixel in the image being water or cloud. Optimal thresholds are empirically determined from a combination of the different bands and their ratios. The thresholds are fixed for reflective bands based on empirical results, whereas thresholds for thermal bands from HLS use the histogram of the image pixel values of the brightness temperatures.

\subsubsection{MNDWI}
Modified normalized difference water index (MNDWI) \cite{xu2006modification} uses the green and SWIR-1 bands from HLS to capture water pixels. Equation~\ref{eq:mndwi} shows the pixel-wise operation to compute the probability of water pixels in the image. SWIR was used due to the observed higher absorption of water in this band \cite{xu2006modification}. At the same time, land tends to reflect SWIR light more than green light, resulting in lower MNDWI for non-water areas.

\begin{equation}
    \text{MNDWI} = \frac{\text{green} - \text{SWIR}}{\text{green} + \text{SWIR}}
    \label{eq:mndwi}
\end{equation}

From the computed values, the final water mask is determined by choosing an optimal static threshold $t$ such that the mask is 1 for pixel values greater than $t$, and 0 otherwise. The threshold is chosen using the validation set. The same threshold is used for evaluation on the test set for comparison with other methods. The Otsu method~\cite{otsu1975threshold} is also explored as a procedure to choose the threshold $t$.

\subsubsection{Deep Learning Methods}
LANA~\cite{zhang2024improvedlana} is a recently released model for cloud and cloud shadow masking. Similar to Fmask, it uses the reflective bands and brightness temperatures to predict cloud masks. It learns cloud masks by utilizing a U-Net architecture (a type of CNN) with attention mechanisms incorporated in the skip connections between the encoder and decoder. Their proposed changes resulted in better performance compared to baseline methods Fmask~\cite{qiu2019fmask} and U-Net Wieland~\cite{wieland2019multi}. In addition to the model, their work also introduced a collection of manually annotated satellite images from Landsat 8 for cloud and cloud shadow prediction. The labels come from USGS personnel annotations, the Spatial Procedures for Automated Removal of Cloud and Shadow (SPARCS) project, and manually annotated tiles from Landsat. Their work resulted in 100 sets of annotated data from different global locations. LANA was trained on 99 out of the 100 sets and evaluated on the remaining set. This procedure was done five times, and the average was obtained and reported as the performance.

U-Net Wieland~\cite{wieland2019multi} was recently introduced in literature for detecting cloud and cloud shadows, and was used as a baseline in LANA~\cite{zhang2024improvedlana}. Similar to our model, it uses the visible, NIR, and SWIR bands. However, the architecture follows U-Net with an encoder, decoder, and skip connections. The model also outputs per-pixel labels to identify clouds and shadows. The model was trained using the specified train and test splits from LANA~\cite{zhang2024improvedlana}, and compared against our model for predicting clouds and cloud shadows.

DeepWaterMap (DWM)~\cite{isikdogan2017surface, isikdogan2019seeing} also uses six bands (visible, NIR, and SWIR bands) as input to predict a water mask. The model adopts a U-Net architecture. However, instead of having large feature maps in the early layers similar to the original U-Net, it was modified to use a constant feature map size throughout the network. Additional changes in the architecture were also applied to save memory such as increasing the stride in the convolution operation instead of using max pooling. For comparison, we train DWM on the DSWx train set for water mask prediction. We use the most updated released version of the model for comparison, while following the training procedure and parameters as closely as possible.

\subsection{Performance Assessment}
The models were evaluated based on (1) masking performance, (2) efficiency, and (3) accuracy of the downstream application to SSC estimation. We discuss each of the evaluation methods below.

\subsubsection{Masking}
The performance for masking is computed through pixel-based metrics F1, recall, precision, and intersection over union (IoU) on the test set. The test set is composed of samples not seen by the model during training and validation. Recall (sometimes referred to as user’s accuracy) is computed as the ratio of correctly predicted pixels to the total number of actual positive pixels in the label as shown in Equation~\ref{eq:recall}. Precision (also called producer’s accuracy) is computed as the ratio of correctly predicted pixels to the total number of pixels that the model predicted as positive as shown in Equation~\ref{eq:precision}. F1 Score in Equation~\ref{eq:f1} is the harmonic mean of recall and precision. Finally, IoU in Equation~\ref{eq:iou} is the ratio of correctly predicted positive pixels to the union of the positive-labeled pixels and the positive-predicted pixels. In the equations below, $TP$ is the true positive pixels, $FN$ is the false negative pixels, and $FP$ is the false positive pixels. For F1 score, recall, precision, and IoU, better performing models would have higher corresponding quantities, with a maximum value of 100\%.

\begin{equation}
    recall = \frac{TP}{TP + FN}
    \label{eq:recall}
\end{equation}
\begin{equation}
    precision = \frac{TP}{TP + FP}
    \label{eq:precision}
\end{equation}
\begin{equation}
    F1\  Score = \frac{2\cdot precision \cdot recall}{precision + recall}
    \label{eq:f1}
\end{equation}

\begin{equation}
    IoU = \frac{TP}{TP + FP + FN}
    \label{eq:iou}
\end{equation}

\begin{figure*}[!t]
    \begin{center}
    \includegraphics[scale=0.41]{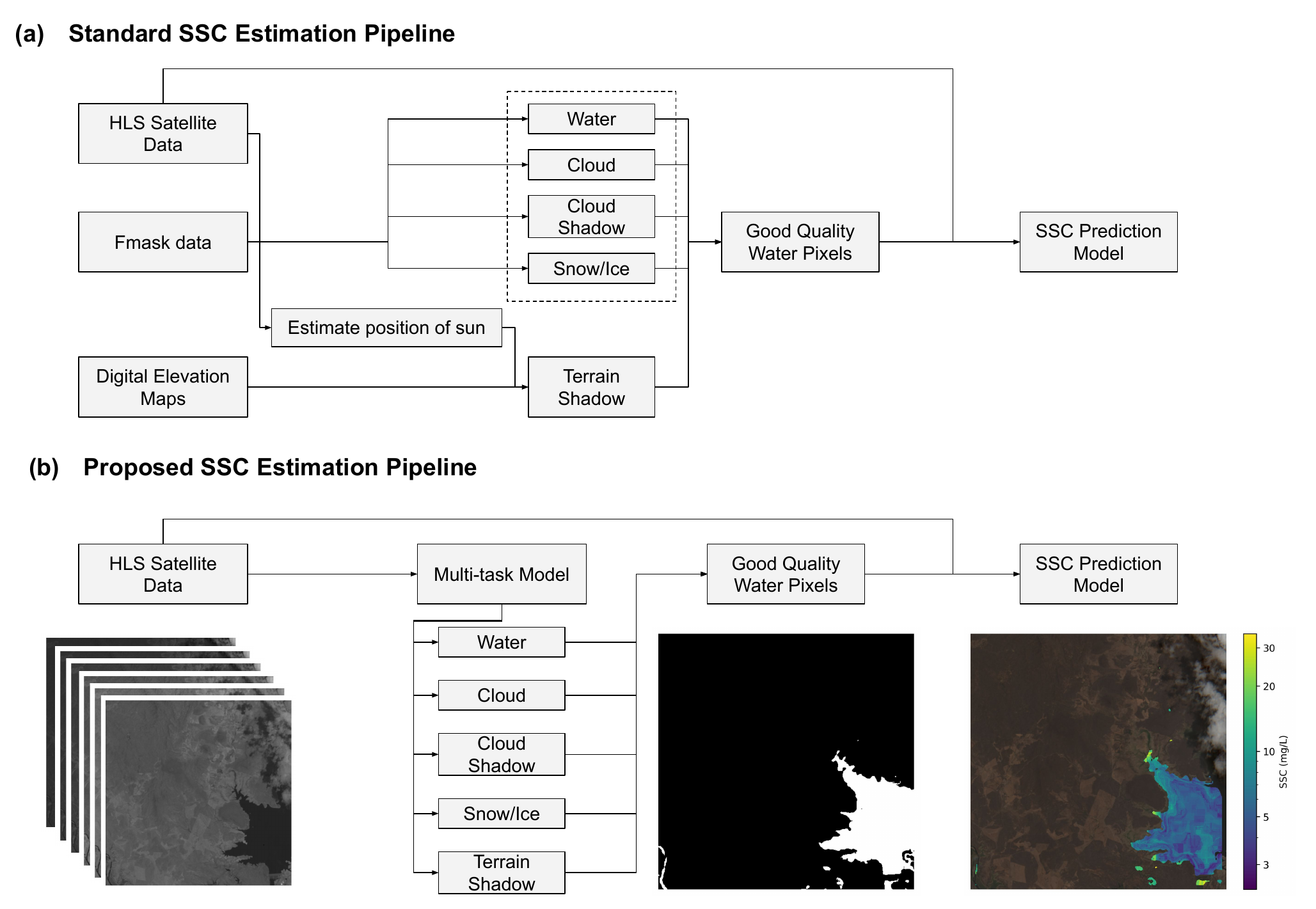}
    \end{center}
    \caption{\textbf{Pipeline for estimating suspended sediment concentration (SSC).} (a) Standard SSC pipelines involve multiple inputs and several processing steps, which contribute to the memory and runtime requirements. (b) Our proposed pipeline only uses readily available Harmonised Landsat-Sentinel (HLS) satellite images and estimates all masks faster by using a single multi-task model. Using good quality water pixels by masking cloud, cloud shadow, snow/ice, and terrain shadow results in significantly better SSC estimates.
    }
    \label{fig:pipeline}
\end{figure*}
\subsubsection{Runtime, Memory, and Storage}
In addition to masking performance, we also evaluate efficiency of the proposed multi-task pipeline. Deploying on a global scale with frequent predictions requires a scalable model in terms of both runtime and memory consumption. We evaluate on both these characteristics, focusing on peak memory consumption and storage costs for the latter. The runtime is measured by aggregating the processing time of the different modules in Fig.~\ref{fig:pipeline}. For comparisons between differerent methods, the runtime does not include downloading HLS data and assumes the HLS is already in memory.

Peak memory consumption would measure the required random access memory (RAM) requirement for running the pipeline. Storage costs are associated with the amount of space (e.g., in hard drive) to store auxiliary data to run the pipeline such as digital elevation maps or pre-trained models. Ideally, the proposed pipeline would require fewer resources (i.e., less total time and less memory) due to the introduction of the multi-task model. The evaluations are done on AMD EPYC 7763 machine where 4 cores and 4GB RAM are used for the process across 100 random samples.

\subsubsection{Effect on Downstream SSC Estimation}

The standard pipeline for SSC estimation from satellite images (Fig.~\ref{fig:pipeline}\revision{a}) takes the HLS tiles, Fmask data, and digital elevation maps as input. Water, cloud, cloud shadow, terrain shadow, and snow/ice masks are estimated to isolate good quality water pixels where the SSC is predicted. The terrain shadow mask is estimated by taking the metadata from HLS tiles to know the date, time, and location of the tile \cite{Soimasuo1994sunmask, michalsky1988astronomical, spencer1971fourier, zimmerman1981sun}. From the metadata, the position of the sun is estimated. Taking the position of the sun and the topography of the area from the DEM, the approximate location of the terrain shadows can be calculated. The other masks required are then obtained as the outputs of Fmask algorithm, which are also accessible from the HLS project. 
\revision{The combination of these masks produce the good quality water pixels.}

\revision{The good quality water pixels are then used as inputs to the SSC prediction model introduced in~\cite{lucchese2024modeling} which uses a two-stage machine learning model. The method uses an ensemble of two neural networks with the first network used for predicting low SSC values (0 to 20 mg/L), and the second network used for predicting middle to high SSC values (14 mg/L and above). Each network is composed of three fully connected layers, following the design of multilayer perceptrons (MLP)~\cite{minsky1969introduction}. Given a location (latitude, longitude) where the SSC is to be estimated, both models take as input the various statistics of the good quality water pixels within 300m of the location. This includes the mean, median, standard deviation, minimum, and maximum for each HLS band. In addition, tiles identified with more than 30\% cloud cover were not used for training. 

The outputs of the two models are then combined based on two empirically determined SSC threshold values using the validation set. That is, if the output of the first model (predicting low SSC values) is below the first threshold, the output of the first model is used. Otherwise, we compare the second model's output (predicting middle to high SSC values) to the second threshold. If the output is above the second threshold, we take the second model’s output. Otherwise, we take the average of the first and second model outputs. The two threshold values that give the lowest error on the validation set are used on the test set for model evaluation. The same threshold values are also used when deploying the model.}

While the standard pipeline is straightforward and uses readily available data, it involves several processing steps and could benefit from not just improving the accuracy of isolating good quality water pixels, but also from streamlining the process. Using a single multi-task model, we can take only HLS data as input and predict multiple masks at the same time. Memory is saved by limiting the input data required to just HLS reflective bands. Run time is also significantly reduced since all the processing is done by a single model. Fig.~\ref{fig:pipeline}\revision{b} shows the proposed pipeline. While we show the pipeline specific to SSC estimation, similar frameworks exist for other applications, and can similarly be optimized by introducing a multi-task masking module.

To further evaluate our framework, we compare the performance of downstream application of SSC predictions when using the standard pipeline without a multi-task model, and the optimized pipeline with the multi-task model. Ideally, we expect the performance of the SSC model to either stay the same or improve with the use of the multi-task model. We use the metrics RMSE (Equation~\ref{eq:rmse}), MAE (Equation~\ref{eq:mae}), and bias (Equation~\ref{eq:bias}) on the predicted SSC values to evaluate the performance of the SSC model across the different pipelines. \revision{We also report various statistics of the absolute error $|y_i - \hat{y}_i|$ for all $i$ in the test dataset---we include the median, maximum, minimum, and standard deviation.} In the following equations, the metrics are computed across $N$ samples, where $y_i$ is the ground truth or SSC label, and $\hat{y}_i$ is the predicted SSC for sample~$i$. An optimal model would have values close to zero for the following metrics.

\begin{equation}
    RMSE = \sqrt{\frac{1}{N}\sum_{i=1}^N (y_i - \hat{y}_i)^2}
    \label{eq:rmse}
\end{equation}

\begin{equation}
    MAE = \frac{1}{N} \sum_{i=1}^N |y_i - \hat{y}_i|
    \label{eq:mae}
\end{equation}

\begin{equation}
    bias = \frac{1}{N} \sum_{i=1}^N \left( y_i - \hat{y}_i \right)
    \label{eq:bias}
\end{equation}

\section{Results}
\label{sec:results}

\subsection{Masking performance}
\subsubsection{Comparison between Multi-task and Single task models}
We compare the quantitative performance of the multi-tasking model to the single-task counterparts. The single-task models use five separate models, where one model is trained on only one task (Fig.~\ref{fig:single-vs-multitask}a). The multi-task model is trained on all five tasks simultaneously (Fig.~\ref{fig:single-vs-multitask}b). Table~\ref{table:segmentation-results} shows that the performance of the multi-task models is either similar or better than the single task models. Three different architectures that use CNN (DeepLabv3+, MobileNetv3) and transformer (Swin-T) were evaluated. Most of the improvement from using multi-task models can be observed on snow/ice masking when using CNN-based models DeepLabv3+ and MobileNetv3 with more than 10\% metric improvement.

\begin{table*}[!t]
    \small
    \begin{center}
    \caption{Masking performance of the multi-task model compared to the single-task models on DSWx test set (highlighted cells indicate better metrics when using the multi-task model compared to its single-task equivalent)
    }
    \label{table:segmentation-results}
    \begin{tabular}{l | c c c c | c c c c}
    \toprule
    \multicolumn{9}{c}{\textbf{DeepLabv3+ (ImageNet pre-trained)}}  \\
    \midrule
    \textbf{Label} & \multicolumn{4}{c|}{\textbf{Single Task Models}} & \multicolumn{4}{c}{\textbf{Multi-task Model}}\\\cline{2-9}
        &   F1 Score  & Precision  &  Recall & IoU &  F1 Score  & Precision  &  Recall & IoU \\
    \midrule
    Water & 89.55\% & 86.99\% & 92.26\% & 81.07\% &         \cellcolor{green!25}89.67\% & \cellcolor{green!25}87.91\% & 91.50\% & \cellcolor{green!25}81.27\% \\
    Cloud Shadow & 61.94\% & 64.39\% & 59.66\% & 44.86\% &  60.16\% & 59.63\% & \cellcolor{green!25}60.69\% & 43.02\% \\
    Cloud & 90.12\% & 89.00\% & 91.27\% & 82.02\% &         87.89\% & \cellcolor{green!25}90.21\% & 85.68\% & 78.39\% \\
    Snow/ice & 60.31\% & 51.81\% & 72.15\% & 43.18\% &      \cellcolor{green!25}70.53\% & \cellcolor{green!25}63.99\% & \cellcolor{green!25}78.56\% & \cellcolor{green!25}54.48\% \\
    Terrain Shadow & 97.33\% & 95.26\% & 99.49\% & 94.79\% & 97.05\% & 94.62\% & \cellcolor{green!25}99.61\% & 94.27\% \\
    \bottomrule
    \multicolumn{9}{c}{\textbf{MobileNetv3 (ImageNet pre-trained)}}  \\
    \midrule
    \textbf{Label} & \multicolumn{4}{c|}{\textbf{Single Task Models}} & \multicolumn{4}{c}{\textbf{Multi-task Model}}\\\cline{2-9}
        &   F1 Score  & Precision  &  Recall & IoU &  F1 Score  & Precision  &  Recall & IoU \\
    \midrule
    Water & 88.03\% & 84.47\% & 91.91\% & 78.63\% &         \cellcolor{green!25}88.18\% & \cellcolor{green!25}85.16\% & 91.42\% & \cellcolor{green!25}78.86\% \\
    Cloud Shadow & 59.44\% & 59.44\% & 59.43\% & 42.29\% &  58.70\% & 58.58\% & 58.82\% & 41.54\% \\
    Cloud & 90.27\% & 90.10\% & 90.44\% & 82.26\% &         89.96\% & \cellcolor{green!25}90.25\% & 89.67\% & 81.75\% \\
    Snow/ice & 59.50\% & 52.78\% & 68.19\% & 42.35\% &      \cellcolor{green!25}74.31\% & \cellcolor{green!25}72.33\% & \cellcolor{green!25}76.41\% & \cellcolor{green!25}59.13\% \\
    Terrain Shadow & 97.30\% & 95.16\% & 99.53\% & 94.74\% & 97.13\% & 94.83\% & \cellcolor{green!25}99.55\% & 94.43\% \\
    \bottomrule
    \multicolumn{9}{c}{\textbf{Swin-T (Satlas pre-trained)}}  \\
    \midrule
    \textbf{Label} & \multicolumn{4}{c|}{\textbf{Single Task Models}} & \multicolumn{4}{c}{\textbf{Multi-task Model}}\\\cline{2-9}
        &   F1 Score  & Precision  &  Recall & IoU &  F1 Score  & Precision  &  Recall & IoU \\
    \midrule
    Water & 89.77\% & 87.35\% & 92.31\% & 81.43\% &         \cellcolor{green!25}91.10\% & \cellcolor{green!25}90.62\% & 91.58\% & \cellcolor{green!25}83.65\% \\
    Cloud Shadow & 64.64\% & 63.78\% & 65.52\% & 47.75\% &  63.10\% & 63.44\% & 62.75\% & 46.09\% \\
    Cloud & 92.59\% & 91.88\% & 93.31\% & 86.20\% &         92.42\% & \cellcolor{green!25}93.14\% & 91.72\% & 85.91\% \\
    Snow/ice & 75.83\% & 72.70\% & 79.23\% & 61.06\% &      \cellcolor{green!25}78.09\% & \cellcolor{green!25}73.92\% & \cellcolor{green!25}82.76\% & \cellcolor{green!25}64.06\% \\
    Terrain Shadow & 97.50\% & 95.70\% & 99.37\% & 95.12\% & 97.26\% & 95.19\% & \cellcolor{green!25}99.44\% & 94.68\% \\
    \bottomrule
    \end{tabular}
    \end{center}
\end{table*}

\subsubsection{Comparison of Multi-task Models Across All Masks}
\begin{figure*}
    \begin{center}
    \includegraphics[scale=0.44]{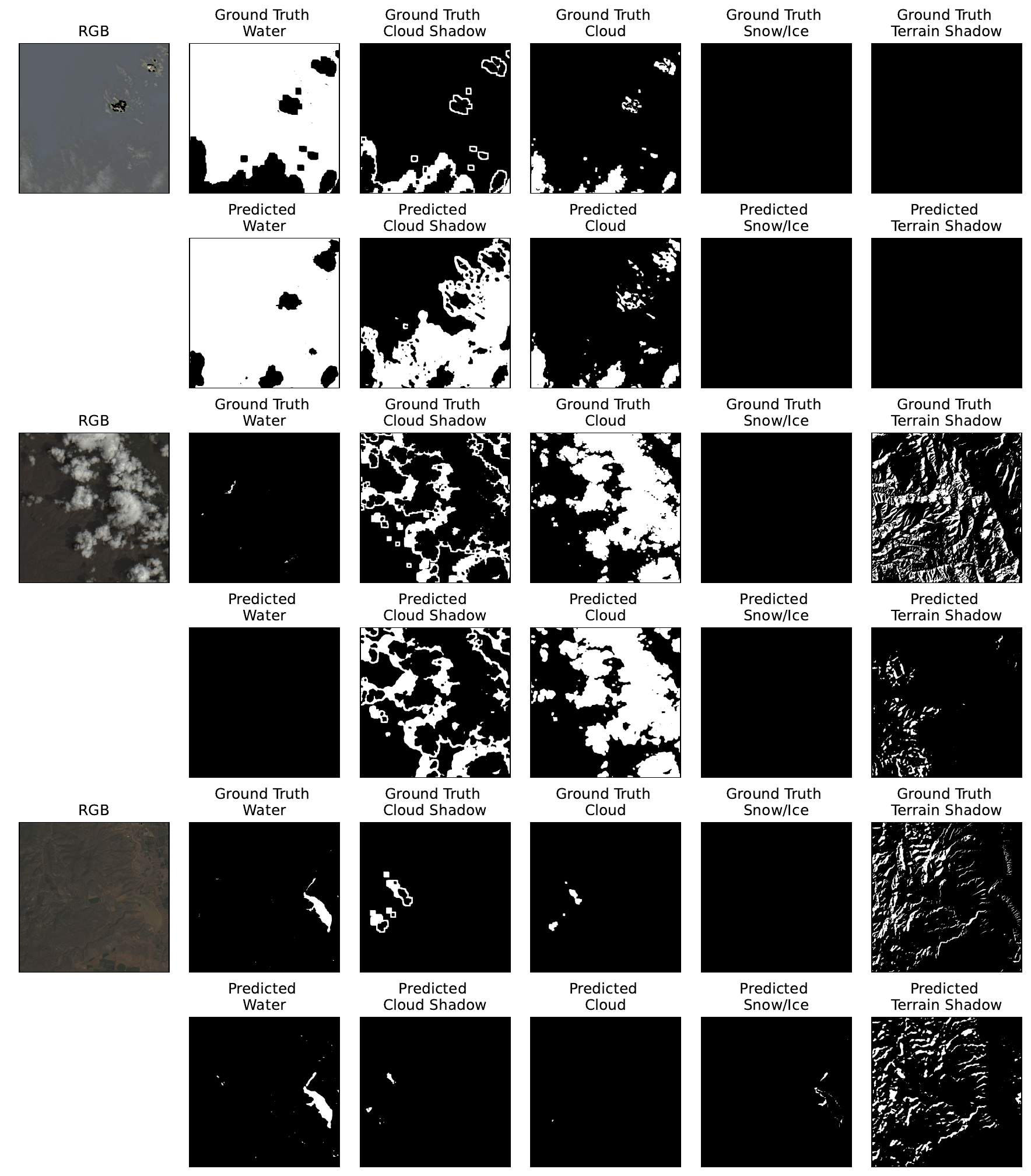}
    \end{center}
    \caption{\textbf{DeepLabv3+ multi-tasking model results on three samples from OPERA DSWx test set.} The RGB images are shown together with the corresponding ground truth and predicted masks. White pixels denote the presence of the mask, and black pixels otherwise. The model predictions across different types of masks closely match the ground truth based on DSWx.
    }
    \label{fig:deeplab-results}
\end{figure*}

All five masks predicted by the multi-task models were evaluated on the held out DSWx test set. Fig.~\ref{fig:deeplab-results} shows qualitative results from training a DeepLabv3+ multi-tasking model across the five masks. The results show DeepLabv3+ can successfully identify water, cloud, cloud shadow, snow/ice, and terrain shadow simultaneously. Table~\ref{table:dswx-f1-score} shows the F1 score across the different masks for various architectures, pre-training methods \revision{(transfer learning)}, and model types. While all multi-task models can reasonably predict the different masks, Swin-T pre-trained on Satlas demonstrates superior performance compared to other architectures. While the architecture itself plays a role in higher accuracy due to its larger capacity and global feature representations, the pre-training method also plays a significant role in improving the performance. When comparing the performance of the same architecture (Swin-T) pre-trained on ImageNet and pre-trained on Satlas, the latter version has as much as 30\% F1 score improvement for cloud shadow masking.  \revision{We additionally find that not applying transfer learning at all reduces performance. For a DeepLabv3+ model that is not pre-trained, water masking on the test set resulted in 86.04\% F1 score, almost a 4\% decrease in performance from the 89.67\% F1 score of the same model pre-trained with ImageNet.}

\begin{table*}[!t]
    \small
    \begin{center}
    \caption{F1 score for various multi-tasking models across mask types on the DSWx test set
    }
    \label{table:dswx-f1-score}
    \setlength{\tabcolsep}{2pt}
    \begin{tabular}{l l l | c c c c c }
    \toprule
    Model & Pre-training & Model Type & Water & Cloud Shadow & Cloud & Snow/Ice & Terrain Shadow \\
    \midrule
    DeepLabv3+ & ImageNet & CNN & \underline{89.67\%} & \underline{60.16\%} & 87.89\% & 70.53\% & 97.05\% \\
    MobileNetv3 & ImageNet & CNN & 88.18\% & 58.70\% & \underline{89.96\%} & \underline{74.31\%} & \underline{97.13\%} \\
    SegNet & ImageNet & CNN & 83.47\% & 58.57\% & 89.36\% & 71.11\% & 97.01\% \\
    ResNet50 & Satlas & CNN & 81.33\% & 53.27\% & 87.92\% & 64.84\% & 96.84\% \\
    Swin-T & Satlas & Transformer & \textbf{91.10\%} & \textbf{63.10\%} & \textbf{92.42\%} & \textbf{78.09\%} & \textbf{97.26\%} \\
    Swin-T & ImageNet & Transformer & 80.73\% & 35.81\% & 85.30\% & 71.36\% & 96.74\% \\
    ViT-B/16 & ImageNet & Transformer & 82.56\% & 37.38\% & 85.89\% & 70.85\% & 96.74\% \\
    ViT-B/16 & Prithvi & Transformer & 76.61\% & 30.87\% & 82.36\% & 69.76\% & 96.74\% \\
    \bottomrule
    \end{tabular}
    \end{center}
    \vspace{3mm}
\end{table*}

\begin{table*}[!t]
    \small
    \begin{center}
    \caption{Performance of the various methods and baselines for water masking
    }
    \label{table:dswx-water-metrics}
    \begin{tabular}{l l l | c c c c }
    \toprule
    Method & Pre-training & Model Type & F1 Score ($\uparrow$) & Precision ($\uparrow$) & Recall ($\uparrow$) & IoU ($\uparrow$) \\
    \midrule
    MNDWI & & & 58.43\% & 78.92\% & 46.39\% & 41.28\% \\
    DWM & & CNN & 82.21\% & 78.54\% & 86.24\% & 69.79\% \\
    \midrule
    DeepLabv3+ & ImageNet & CNN & \underline{89.67\%} & \underline{87.91\%} & \underline{91.50\%} & \underline{81.27\%} \\
    MobileNetv3 & ImageNet & CNN & 88.18\% & 85.16\% & 91.42\% & 78.86\% \\
    SegNet & ImageNet & CNN & 83.47\% & 82.94\% & 84.01\% & 71.63\% \\
    ResNet50 & Satlas & CNN & 81.33\% & 78.76\% & 84.08\% & 68.54\% \\
    Swin-T & Satlas & Transformer & \textbf{91.10\%} & \textbf{90.62\%} & \textbf{91.58\%} & \textbf{83.65\%} \\
    Swin-T & ImageNet & Transformer & 80.73\% & 77.88\% & 83.80\% & 67.69\% \\
    ViT-B/16 & ImageNet & Transformer & 82.56\% & 81.15\% & 84.03\% & 70.30\% \\
    ViT-B/16 & Prithvi & Transformer & 76.61\% & 74.60\% & 78.74\% & 62.09\% \\
    \bottomrule
    \end{tabular}
    \end{center}
\end{table*}

\begin{figure*}
    \begin{center}
    \includegraphics[scale=0.38]{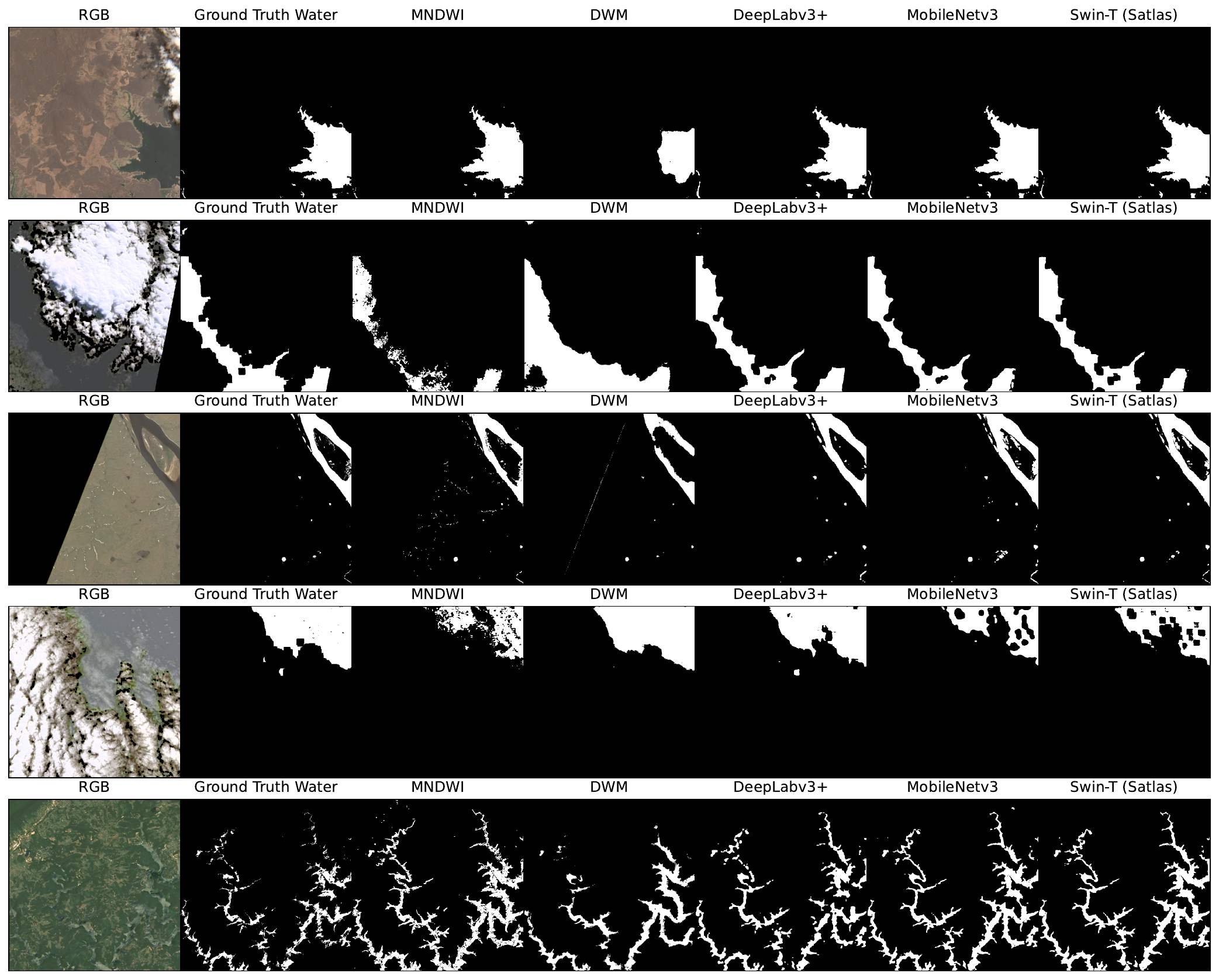}
    \end{center}
    \caption{\textbf{Water masking results of different methods.} MNDWI results shown here were filtered to remove clouds and shadows, while results from other methods were not filtered in any way. MNDWI fails to identify several water pixels, while DWM tends to predict smooth water boundaries which could miss details. DeepLabv3+, MobileNetv3, and Swin-T outperform the baselines but DeepLabv3+ is more robust to noise (see row four in the figure).
    }
    \label{fig:water-results}
\end{figure*} 
\subsubsection{Water Mask comparison of Multi-task Models with Baselines}
In addition to the performance of simultaneously predicting all masks, the performances of individual masks were also compared and evaluated. Table~\ref{table:dswx-water-metrics} shows the performance of the different methods for water masking using the DSWx test set. In the evaluation, only the water mask output of the multi-task model was used; other masks were discarded. The baseline method DWM~\cite{isikdogan2019seeing} was trained on the train set of DSWx and only predicts water masks. It was trained with the same parameters and setup recommended in their paper. The baseline MNDWI was evaluated out of the box for fair comparison, resulting in a 16.62\% F1 score. However, standard practice in hydrology has moved beyond straightforward application of MNDWI. To achieve results similar in practice, we apply cloud and shadow masking from DSWx and remove predictions over regions with no data. We include these results in Table~\ref{table:dswx-water-metrics}, resulting in a 58.43\% F1 score. Other methods were not applied with the same cloud and shadow masking, yet the results still show multi-task models Swin-T (Satlas pre-trained) and DeepLabv3+ (ImageNet pre-trained) have the best performance when masking water pixels with 91.10\% and 89.67\% F1 score, respectively. MobileNetv3 also has competitive performance against other models. The multi-task models introduce around 10\% F1 score improvement compared to baseline methods DWM and MNDWI.

Fig.~\ref{fig:water-results} shows results of predicting the water mask across different methods. The top 3 best performing multi-task models are displayed and compared against the baselines. The visual results show that MNDWI is sensitive high reflectance values, which leads to pixels that are falsely identified as water. It also frequently fails to identify water pixels. DWM tends to predict smooth water boundaries, and can fail to capture fine details such as branching out in rivers. Due to the multi-tasking setup of the proposed model, the outputs are less sensitive to clouds and shadows, since the model is trained to identify all the different types of masks at the same time. At the same time, the multi-task models are able to capture fine details in the water boundaries.

\subsubsection{Cloud Mask comparison of Multi-task Models with Baselines}
The cloud masking performance of the multi-task models were also evaluated on both the DSWx test set, and the LANA~\cite{zhang2024improvedlana} benchmark. The evaluation on the LANA benchmark involved training the multi-task models on the sets of data not used for evaluation, and evaluated on five unseen sets of data, similar to the evaluation outlined in their paper for fair comparison. The dataset contains manually labeled cloud pixels from USGS personnel. Table~\ref{table:lana-metrics} shows the difference in performance between the different multi-task models and the baselines LANA, Fmask, and U-Net Wieland. 
\revision{The baselines are models developed specifically for cloud and cloud shadow masking.}
Although the proposed multi-task models are using only 6 bands from the Landsat data -- as opposed to LANA and Fmask that use 8 bands for prediction -- the performance is similar (for cloud masking), or even better (for cloud shadow and clear masks) than the baselines. 

\revision{As additional reference, a previous method of cloud and cloud shadow detection on Landsat data~\cite{chai2019cloud} explored different CNN models and compared against a variant of Fmask (CFmask). They obtained around 94\% overall accuracy for detecting cloud-based classes (cloud, cloud shadow, thin cloud, clear), and improved over CFmask accuracy by around 10\%. However, their sampling of Landsat data into training, validation, and test sets differs from our setup since they sample their data at the patch-level, where each training patch is a small crop of the larger Landsat image scene (and could potentially have an overlap with another patch in the validation/test set), similar to \cite{wang2021all, hu2021cdunet, yao2023cloud}. In contrast, our training uses data sampled at the scene-level, ensuring that training and testing patches do not come from the same image and have no spatial overlap, similar to LANA [9]. We do this to avoid inflating the reported metrics since patches from the same scene could have similar cloud and surface conditions.}

\begin{table*}[!t]
    \small
    \begin{center}
    
    \caption{Performance of cloud masking on manually labeled LANA~\cite{zhang2024improvedlana} dataset
    }
    \label{table:lana-metrics}
    \begin{tabular}{l l | c c c }
    \toprule
    & Model & Cloud ($\uparrow$) & Cloud Shadow ($\uparrow$) & Clear ($\uparrow$) \\
    \midrule
    Baselines & LANA \cite{zhang2024improvedlana} & 92.42\% & 57.53\% & 89.02\% \\
        & Fmask \cite{qiu2019fmask} & 89.81\% & 45.42\% & 88.09\% \\
        & U-Net Wieland \cite{wieland2019multi} & 87.68\% & 52.06\% & 86.19\% \\
    \midrule
    Multi-task Models & DeepLabv3+ (ImageNet pre-trained) & 92.64\% & \underline{65.79\%} & 95.54\% \\
        & MobileNetv3 (ImageNet pre-trained) & \textbf{93.70\%} & 63.60\% & \underline{95.77\%} \\
        & SegNet (ImageNet pre-trained) & 91.19\% & 57.64\% & 95.19\% \\
        & ResNet50 (Satlas pre-trained) & 85.78\% & 63.67\% & 92.77\% \\
        & Swin-T (Satlas pre-trained) & \underline{92.96\%} & \textbf{69.56\%} & \textbf{95.80\%} \\
        & Swin-T (ImageNet pre-trained) & 82.73\% & 4.32\% & 92.49\% \\
        & Vit-B/16 (ImageNet pre-trained) & 59.89\% & 0.01\% & 88.16\% \\
        & Vit-B/16 (Prithvi pre-trained) & 81.38\% & 6.94\% & 91.52\% \\
    \bottomrule
    \end{tabular}
    \end{center}
\end{table*}

\begin{table*}[!t]
    \small
    \begin{center}
    
    \caption{Runtime comparison of standard SSC pipeline and the proposed multi-task SSC pipeline on 4 cores of an AMD EPYC 7763 machine
    }
    \label{table:runtime-results}
    \begin{tabular}{l | c c c }
    \toprule
    & Runtime on  & Runtime on  & Improvement \\
    & 1 sample (s) & 400k samples (days) & (\%)\\
    \midrule
    Standard SSC Pipeline & 18.757 & 86.84 & - \\
    \midrule
    DeepLabv3+ (ImageNet pre-trained) & 2.002 & 9.27 & 89.33\% \\
    MobileNetv3 (ImageNet pre-trained) & \textbf{0.601} & \textbf{2.78} & \textbf{96.80\%} \\
    SegNet (ImageNet pre-trained) & 2.259 & 10.46 & 87.96\% \\
    ResNet50 (Satlas pre-trained) & 7.209 & 33.38 & 61.57\% \\
    SwinT (Satlas pre-trained) & 6.260 & 28.98 & 66.62\% \\
    SwinT (ImageNet pre-trained) & \underline{1.254} & \underline{5.80} & \underline{93.32\%} \\
    ViT-B/16 (ImageNet pre-trained) & 2.450 & 11.34 & 86.94\% \\
    ViT-B/16 (Prithvi pre-trained) & 3.493 & 16.17 & 81.38\% \\
    \bottomrule
    \end{tabular}
    \end{center}
\end{table*}

\begin{figure*}
    \begin{center}
    \includegraphics[scale=0.45]{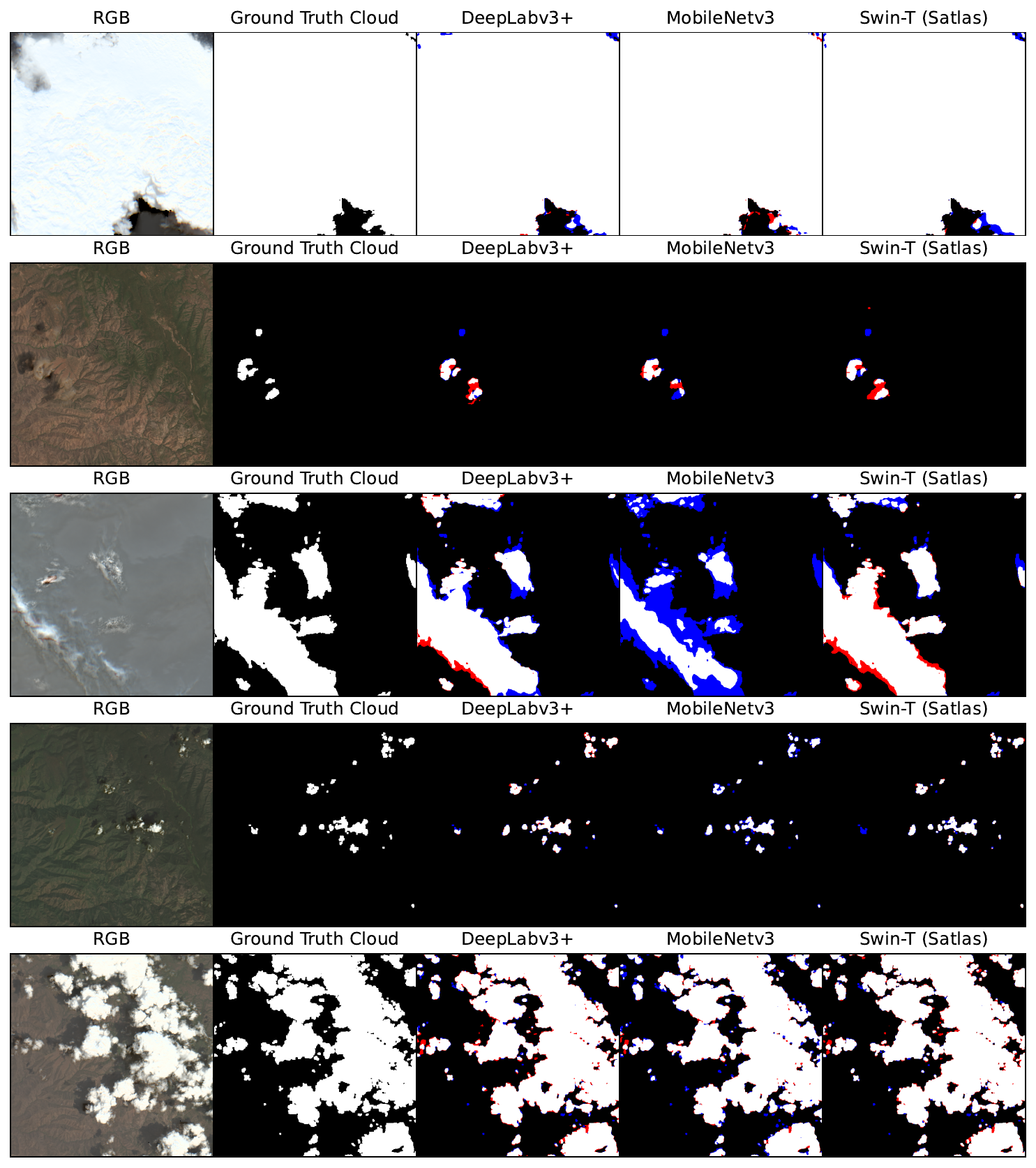}
    \end{center}
    \caption{\textbf{Cloud masking results for different methods on the test set.} False positive pixels are colored red, and false negatives are colored blue. While all models perform similarly, MobileNetv3 tends to miss more true positives, i.e., it incorrectly classifies could pixels as non-cloud.
    }
    \label{fig:dswx-cloud-results}
\end{figure*}

\begin{table*}
    \begin{center}
    \caption{Runtime breakdown of a standard SSC pipeline and our proposed multi-task SSC pipeline with DeepLabv3+ on 4 cores of an AMD EPYC 7763 machine for a single sample
    }
    \label{table:runtime-breakdown}
    \begin{tabular}{l|l|l|l|l}
    \toprule
                    & Standard SSC Pipeline     & Time (s)              & Proposed SSC Pipeline     & Time (s)              \\
    \midrule
    Pre-processing  & Estimate position of sun  & 2.05                  & -                         & -                     \\
    \midrule
    \multirow{5}{*}{Obtaining Masks} & Water                     & \multirow{4}{*}{1.94} & Water                     & \multirow{5}{*}{1.96} \\
                    & Cloud                     &                       & Cloud                     &                       \\
                    & Cloud Shadow              &                       & Cloud Shadow              &                       \\
                    & Snow/ice                  &                       & Snow/ice                  &                       \\
                    \cmidrule{2-3}
                    & Terrain Shadow            & 5.61                  & Terrain Shadow            &                       \\
    \midrule
    Combining Masks & Good Quality Water Pixels & 9.10                  & Good Quality Water Pixels & 0.04      \\           
    \bottomrule
    \end{tabular}
    \end{center}
\end{table*}

\begin{figure*}
    \begin{center}
    \includegraphics[scale=0.44]{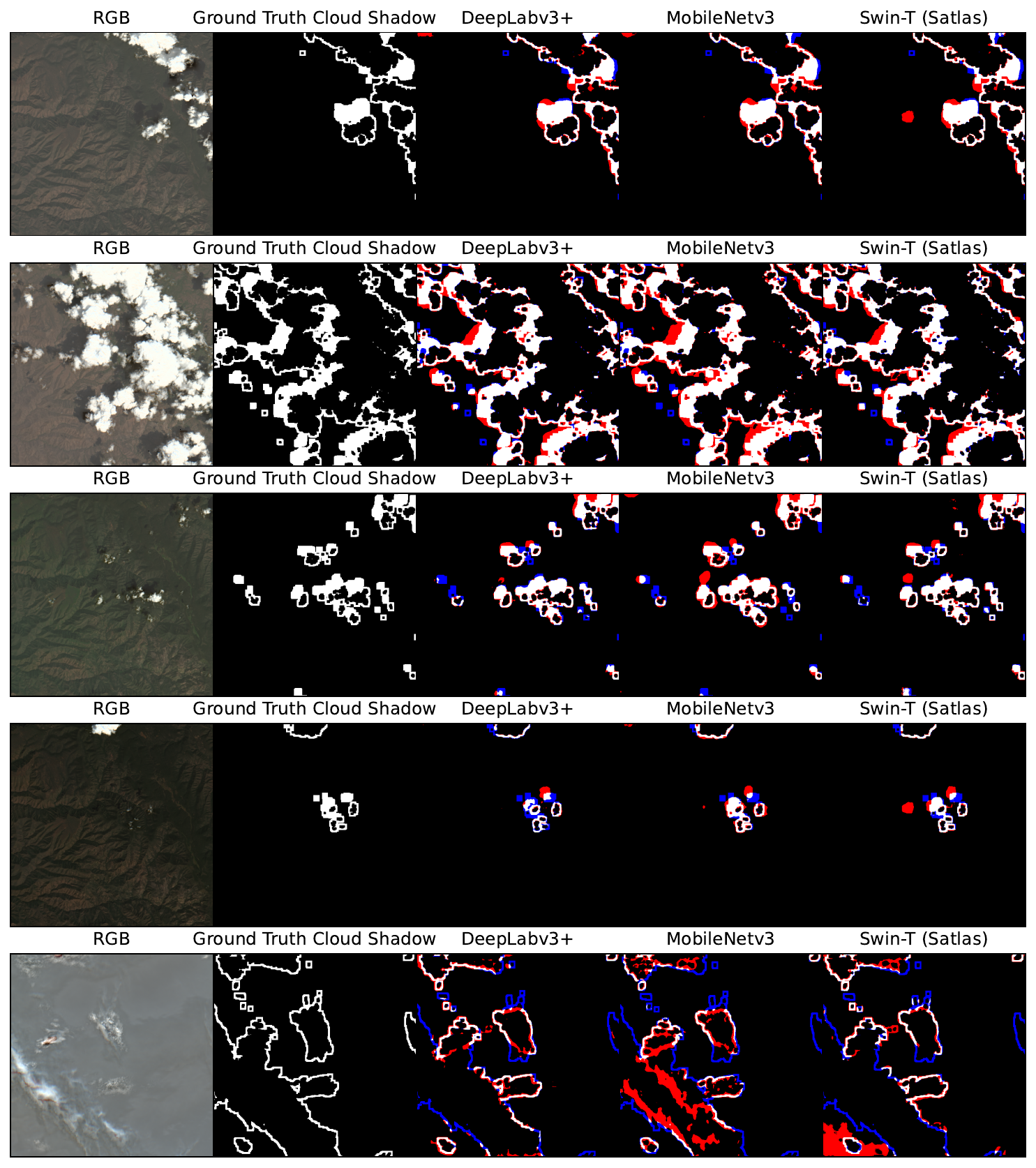}
    \end{center}
    \caption{\textbf{Cloud shadow masking results for different methods on the test set.} False positive pixels are colored red, and false negatives are colored blue. The performance of the models on cloud shadows is similar to cloud masking, and show that the models have comparable results.
    }
    \label{fig:dswx-cloudshadow-results}
\end{figure*}

Figures~\ref{fig:dswx-cloud-results} and \ref{fig:dswx-cloudshadow-results} show additional results for predicting cloud and cloud shadow pixels of the multi-task models on the DSWx test set. These model results were based on training on the DSWx train set. The results show that all three models DeepLabv3+, MobileNetv3, and Swin-T can accurately predict both clouds and cloud shadows when compared to the ground truth labels, as supported by the results in Table~\ref{table:dswx-f1-score}. There are slight differences in MobileNetv3 and DeepLabv3+, where MobileNetv3 would sometimes miss true positive pixels, and DeepLabv3+ would sometimes have false positive pixels. However, overall, the performance of the models are comparable.

\subsubsection{Additional Qualitative Results on Other Masks}
\begin{figure*}
    \begin{center}
    \includegraphics[scale=0.45]{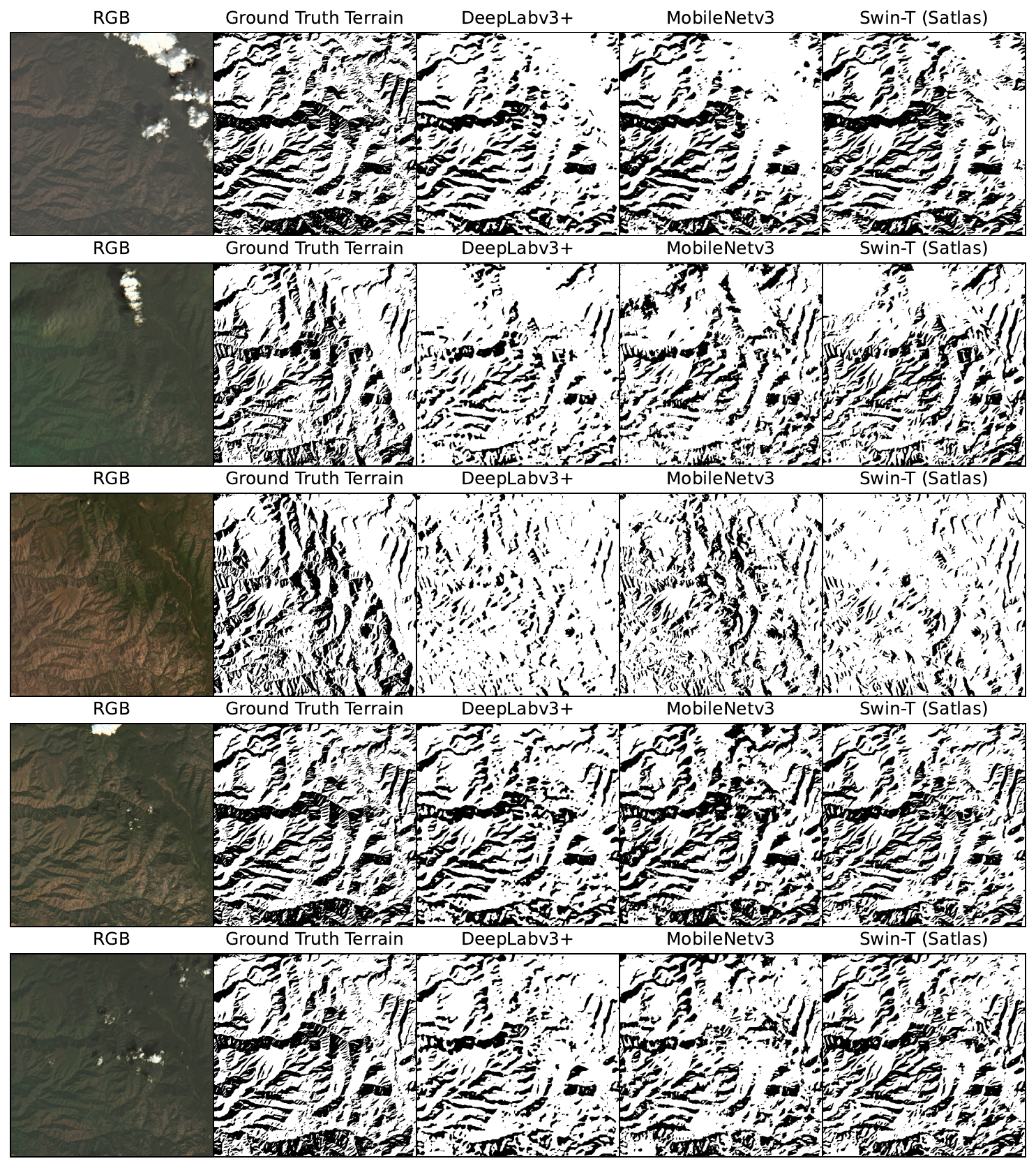}
    \end{center}
    \caption{\textbf{Terrain shadow masking results of different methods on the test set.}  DeepLabv3+ is more sensitive to clouds and its surrounding pixels, while MobileNetv3 and Swin-T can still predict near clouds.
    }
    \label{fig:dswx-terrain-results}
\end{figure*}

\begin{figure*}
    \begin{center}
    \includegraphics[scale=0.45]{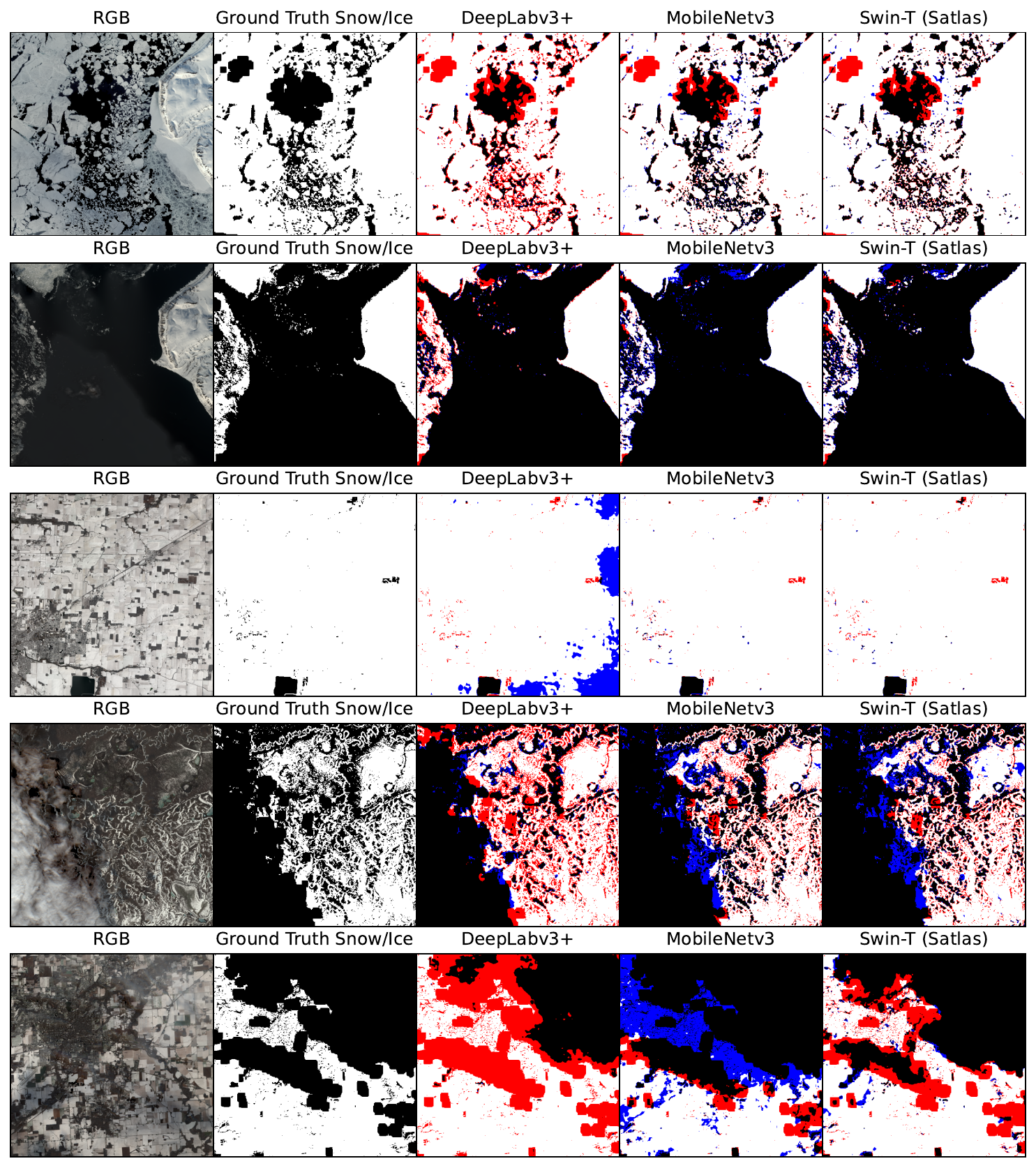}
    \end{center}
    \caption{\textbf{Snow/ice masking results of different methods on the test set.} Results from MobileNetv3 and Swin-T are able to capture true positive pixels for snow/ice. While DeepLabv3+ classifies more pixels as snow/ice, there are more false positives (red pixels) present. On the other hand, MobileNetv3 has more false negatives (blue pixels).
    }
    \label{fig:dswx-snowice-results}
\end{figure*}

Fig.~\ref{fig:dswx-terrain-results} and \ref{fig:dswx-snowice-results} show qualitative results on predicting terrain shadow and snow/ice, respectively. The three best performing multi-task models are visualized with the RGB bands and the label for each of the samples. Looking at the visual results in Fig.~\ref{fig:dswx-terrain-results} for terrain shadow, Swin-T pre-trained on Satlas have closely aligned outputs with the ground truth, where it's able to capture finer details and even predict parts partially occluded by clouds. DeepLabv3+ terrain predictions are more sensitive to cloud occlusions in comparison. This supports the reported F1 score in Table~\ref{table:dswx-f1-score} for terrain shadow where Swin-T slightly outperforms DeepLabv3+ and MobileNetv3.

Snow/ice pixel identification results in Fig.~\ref{fig:dswx-snowice-results} show DeepLabv3+ being able to identify more snow/ice pixels, but it tends to exceed the boundaries and falsely predict surrounding pixels as snow/ice, and thus has more false positives. MobileNetv3 and Swin-T have less false positives and while the predictions show less pixels identified as snow/ice, the positive predictions appear in accurate locations. As a result, MobileNetv3 and Swin-T have higher quantitative metrics on snow/ice masking (Table~\ref{table:dswx-f1-score}).

\begin{table*}[!t]
    \small
    \begin{center}
    
    \caption{Peak memory consumption and storage overhead comparison between the standard and proposed multi-task SSC pipelines
    }
    \label{table:memory-results}
    \begin{tabular}{l | c c}
    \toprule
    & Peak Memory & Storage \\
    & Consumption (GB) & Overhead (GB) \\
    \midrule
    Standard SSC Pipeline & 1.851 & 96.000 \\
    \midrule
    DeepLabv3+ (ImageNet pre-trained) & 1.589 & 0.257 \\
    MobileNetv3 (ImageNet pre-trained) & \textbf{0.833} & \textbf{0.014} \\
    SegNet (ImageNet pre-trained) & 1.311 & 0.288 \\
    ResNet50 (Satlas pre-trained) & 1.179 & 0.304 \\
    Swin-T (Satlas pre-trained) & 1.290 & 0.343 \\
    Swin-T (ImageNet pre-trained) & 1.235 & 0.361 \\
    ViT-B/16 (ImageNet pre-trained) & 2.072 & 1.000 \\
    ViT-B/16 (Prithvi pre-trained) & 2.625 & 1.260 \\
    \bottomrule
    \end{tabular}
    \end{center}
\end{table*}

\subsection{Effect on Downstream Application: SSC Estimation}
\subsubsection{Time}

Table~\ref{table:runtime-results} shows comparisons on the amount of time to process a single sample for an SSC model, and the amount of time to process 400k samples. On average, there are 400k samples per day released as part of HLS based on the data from EOSDIS. Assuming a linearly increasing runtime based on a single sample, processing 400k samples would take approximately 86.84 days when using the standard SSC pipeline. In contrast, using a MobileNetv3 multi-task model would result in only 2.78 days for 400k samples, a 30$\times$ speedup (or 96.80\% runtime improvement). Table~\ref{table:runtime-breakdown} shows how each module contributes to the amount of time to process a single sample. Majority of the runtime in a standard SSC pipeline comes from combining masks from different sources to get good quality water pixels. Unlike the standard SSC pipeline that requires reprojections and alignments due to the different sources of data, our proposed pipeline predicts masks that can be directly combined without additional processing overhead. Four cores of an AMD EPYC 7763 was used to evaluate the runtime and memory.

\subsubsection{Memory and Storage}
Table~\ref{table:memory-results} shows the difference in memory consumption between the standard and the proposed multi-task pipelines. The peak memory consumption is measured by running the pipelines to produce features for the SSC model. The standard SSC pipeline requires at least 1.85~GB of RAM. However, when using MobileNetv3 as the multi-task model in the proposed pipeline, the required RAM reduces to 0.833~GB, less than half the requirement of the standard pipeline. The other models, DeepLabv3+ and Swin-T, also require less RAM than the standard pipeline. At the same time, the storage costs associated with any of the multi-task models are about a hundredth of the standard SSC pipeline, since the proposed pipeline would only need to store the pre-trained models. The standard pipeline requires storing all digital elevation maps covering all landmasses, which is 96~GB in size, while the proposed multi-task pipeline only requires the models themselves to be stored, which is around 1~GB in size.

\subsubsection{Accuracy}
The effect of using a more accurate multi-task model was also evaluated on the downstream application of SSC estimation. DeepLabv3+ pre-trained on ImageNet was used as the multi-task model. While MobileNetv3 is faster, DeepLabv3+ showed better performance for water masking and is smaller than Swin-T pre-trained on Satlas. Table~\ref{table:ssc-results} shows the performance of the SSC model when trained on features produced by the standard SSC pipeline using Fmask and DEM compared to the SSC model when trained on features produced by the optimized multi-task pipeline. \revision{We report the statistics of the error on the test set of the dataset.} The model from \cite{lucchese2024modeling} for SSC estimation was trained and evaluated as outlined in their methodology. The results show that the performance of the SSC model either improved or stayed similar across the metrics when using the proposed multi-task pipeline. The RMSE between ground truth and predicted SSC values reduced by 2.64 mg/L. \revision{There is also significant reduction in the 95th (E95) and 90th (E90) percentile error with 13.55 mg/L and 6.25mg/L error reduction, respectively.} This can be attributed to better quality water pixels isolated for training the SSC model, as produced by the multi-task model. Fmask, the water mask used in the standard SSC pipeline, tends to falsely identify areas of water which could contribute to incorrect features being used to estimate SSC values. 

 \begin{table}[t]
     \small
     \begin{center}
     \revision{
     \caption{SSC model prediction metrics using the standard SSC pipeline and the proposed multi-task pipeline using a two-stage ML model ($\downarrow$ means lower is better)
     }
     \vspace{-3mm}
     \label{table:ssc-results}
      \setlength{\tabcolsep}{3pt}

    \begin{tabular}{l | r r | r}
    \toprule
    \multicolumn{4}{c}{\textbf{SSC Prediction Metrics} (mg/L)}  \\
    \midrule
    \multirow{2}{*}{Metric} & Standard & Multi-task & \multirow{2}{*}{$\Delta (\downarrow)$} \\
     & SSC Pipeline & SSC Pipeline & \\
    \midrule
    RMSE ($\downarrow$)       & 60.61 & 57.97 &       - 2.64 \\
    MAE ($\downarrow$)        & 24.44 & 22.88 &       - 1.56 \\
    Bias ($\downarrow$)       & 22.53 & 21.23 &       - 1.30 \\
    Max Abs Error ($\downarrow$)        & 701.05 & 700.40 &     - 0.65 \\
    E95 Abs Error ($\downarrow$)        & 119.22 & 105.67 &     - 13.55 \\
    E90 Abs Error ($\downarrow$)        & 63.86 & 57.61 &       - 6.25 \\
    E75 Abs Error ($\downarrow$)        & 18.26 & 16.60 &       - 1.66 \\
    Median Abs Error ($\downarrow$)     & 5.73 & 5.34 &         - 0.39 \\
    Std Dev Abs Error ($\downarrow$)    & 55.96 & 52.97 &       - 2.99 \\
    Min Abs Error ($\downarrow$)        & 0.00 & 0.02 &         + 0.02 \\
    \bottomrule
    \end{tabular}
    
     }
     \vspace{-6mm}
     \end{center}
 \end{table}

\section{Discussion}
\label{sec:discussion}
Classifying pixels from satellite images for downstream applications on a global level requires both accurate and efficient frameworks. Evaluation should also be done on the full end-to-end pipeline to capture the positive effects, if any, of the introduced masking. SSC estimation for all rivers in all landmasses, for example, relies on good quality water pixels not occluded by clouds, cloud shadows, terrain shadows, and snow/ice. At the same time, frequent analysis of SSC in global surface waters would only be possible with pipelines that produce these masks at a fast rate. A standard SSC pipeline uses the Fmask algorithm on HLS and digital elevation maps to isolate good quality water pixels, but the reprojections and alignments needed to process information from different data sources introduce compute overhead. In addition, Fmask has limitations for accurately isolating water pixels due to its dependence on empirically-derived thresholds. We introduced a more efficient framework that uses a multi-task model to predict all necessary masks for SSC estimation while illustrating that its improved masking performance also results in better performance for SSC estimation.

\subsection{Masking Performance}
We showed through our experiments that the multi-task model can benefit from learning different masks in a consistent manner. In the DSWx dataset, since all five masks are predicted together, the multi-task model has context on where clouds are, and can learn to predict other masks such as water and terrain shadow that are not occluded by clouds. This advantage is apparent for the water and snow/ice masks that show significant improvement when trained with the multi-task model over the single task model equivalents (Table~\ref{table:segmentation-results}). There are especially large improvements for snow/ice in particular for CNN models DeepLabv3+ and MobileNetv3 with more than 10\% improvement on all metrics. This could be due to the smaller number of positive samples for snow/ice globally which limits the accuracy of the single task model. In contrast, the ability of the multi-task model to identify other types of masks (e.g., water) can significantly help the model identify which ones are not snow/ice and reduce false positives as observed in the increase in precision for the multi-task models. The advantage is further supported by the higher accuracy in both water mask (Table~\ref{table:dswx-water-metrics}, Fig.~\ref{fig:water-results}) and cloud mask (Table~\ref{table:lana-metrics}) experiments compared to existing models LANA, Fmask, U-Net Wieland, and DWM.

Performance of both single-task and multi-task models are relatively lower for snow/ice and cloud shadow masks in DSWx compared to other masks such as water and cloud masks (Table~\ref{table:segmentation-results}). This is expected due to the smaller number of positive samples the models can learn from as reported in Table~\ref{table:data-ave-pixels}. These classes are also harder to classify due to ambiguity with other classes, such as clouds that could be mistaken for cloud shadow and vice versa. Similar phenomena could also be observed for snow/ice and clouds. In contrast, there are a large number of positive samples for cloud and terrain shadow, where we see better performance across all models (around 90\% F1 score). In addition to limited snow/ice samples that the model can learn from, it should also be noted that while DSWx provides sufficient labels for snow/ice, there are occasional misclassifications. These typically occur over waters with unusual colors due to high sediment concentrations or dissolved solids (Jones, 2019). It is possible that some disagreements between model predictions and labels come from erroneous labels. Examining manual cloud annotations from LANA, and comparing them against DSWx labels, we find that cloud labels from DSWx have an F1 score of 89.81\%.

Experiments on the LANA dataset (Table~\ref{table:lana-metrics}) show that predicting cloud shadow is more challenging than clouds or clear pixels due to the smaller number of samples available. Clouds and clear pixels are also easier to identify due to their distinct features. However, there is at least a 10\% F1 score improvement in the prediction of cloud shadow when using the multi-task model Swin-T (pre-trained with Satlas) when compared with LANA, Fmask, or U-Net Wieland. Similar to previous results, DeepLabv3+ and MobileNetv3 also show competitive results with Swin-T. While Swin-T performs well, other transformer architectures pre-trained with ImageNet or Prithvi have significantly lower performance on the LANA dataset as compared to the DSWx dataset. This is due to the small number of data samples used for training; the LANA dataset only contains around 16,000 training samples as opposed to the DSWx dataset with more than 80,000 training samples.

Although DSWx labels are also generated by an algorithm, DSWx requires data from Copernicus DEM, Copernicus land cover, ESA worldcover, NOAA GSHHS shapefile, and HLS~\cite{jones2019improved}. The workflow for generating DSWx labels involve applying multiple steps to the aforementioned data sources such as filtering, and conducting diagnostic tests to produce masks. While their method is sufficiently robust for detecting water, clouds, and shadows, applying it on large amounts of data (e.g., decades of historical data from Landsat and Sentinel) would be a difficult task. Our work introduces a promising framework that runs fast using only HLS data, making it possible to run even on past satellite images. In addition, we show that our framework can be used on manually labeled data from LANA~\cite{zhang2024improvedlana} (Table~\ref{table:lana-metrics}). Even when evaluating on datasets with manually labeled data, we show our method's generated masks perform better than current state-of-the-art methods. These results are encouraging and offer an interesting way forward for operational production of OPERA data: the huge effort required to generate DSWx can be used and honored with a more efficient and likely more accurate (via LANA) representation of itself, potentially saving resources for the operational production of OPERA.

\subsection{SSC estimation end-to-end pipeline efficiency}
The standard SSC pipeline had a measured processing of 86.84 days, which is impractical when running almost daily SSC predictions for in-depth global surface water analyses (Table~\ref{table:runtime-results}). As an alternative, multiple machines can be used in parallel to reduce processing time. However, this would require significant resources and result in larger costs. The large memory overhead required for a standard pipeline (Table~\ref{table:memory-results}) also contribute to the impracticality of increasing the hardware scale to reduce processing time. Table~\ref{table:runtime-breakdown} shows the breakdown of the runtime and how different modules contribute to the processing time. The bulk of the processing time comes from aligning the different masks to isolate good quality pixels. Since different masks come from different sources (e.g., DEM or Fmask), the projection could be different, requiring warping and reprojections to be applied.

With the proposed multi-task pipeline, all the reprojections that come from using different data sources are eliminated. The processing time can be reduced by as much as 96.80\%, such that 400k samples can be processed in less than 3 days on a 4-core CPU (Table~\ref{table:runtime-results}). This approach would require fewer machines in parallel, if necessary at all. MobileNetv3+ has the fastest runtime of 2.78 days for 400k samples, with Swin-T pre-trained with ImageNet and DeepLabv3+ having runtimes of 5.80 and 9.27 days, respectively.

The multi-task pipeline also has comparably lower requirements for memory and storage compared to the standard pipeline (Table~\ref{table:memory-results}). When running multiple parallel processes for obtaining features for SSC models, this could significantly reduce costs. In addition to efficiency improvements, it was shown that our proposed multi-task pipeline improves SSC estimation accuracy (Table~\ref{table:ssc-results}). Taking the improvements in performance, runtime, and memory consumption, the multi-task pipeline presents a promising framework for downstream applications such as global SSC prediction. 

\subsection{Performance of different architectures and pre-training}
Experiments discussed in this work show that using multi-task models for simultaneously predicting water, cloud, cloud shadow, terrain shadow, snow/ice result in better performance over its single task equivalent while requiring less training resources (Fig.~\ref{fig:single-vs-multitask}). Different architectures MobileNetv3, DeepLabv3+, and Swin-T also show larger performance improvements when using a multi-task model for predicting labels with small positive samples (e.g., snow/ice). Our experiments have shown that a multi-task model using Swin-T pre-trained with Satlas have superior masking accuracy compared to baselines and other multi-task models using a different architecture. This was illustrated on different types of masks. For water masking, Swin-T performed better than Fmask~\cite{qiu2019fmask}, MNDWI~\cite{xu2006modification}, and DWM~\cite{isikdogan2019seeing} with at least a 10\% increase in F1 score. Swin-T also outperformed other baselines Fmask, LANA, and U-Net Wieland on cloud masking on the LANA-introduced dataset, where Swin-T has 10\%, 12\%, and 6\% F1 score improvement for cloud, cloud shadow, and clear pixel identification, respectively (Table~\ref{table:lana-metrics}). More generally, Swin-T pre-trained on Satlas was also shown to perform better than other architectures for predicting five masks simultaneously (Table~\ref{table:dswx-f1-score}). 

\revision{Our experiments also show the effect of transfer learning through various pre-training datasets.} CNN models DeepLabv3+ and MobileNetv3, both ImageNet pre-trained, show competitive masking performance when compared to Swin-T, with only around 3\% F1 score difference across most masks (Table~\ref{table:dswx-f1-score}). Despite also being pre-trained on ImageNet, transformer models Swin-T and ViT-B/16 have lower F1 score \revision{than} their CNN counterparts. This could be attributed to the lack of inductive bias in transformers, requiring models to be trained with larger datasets (e.g., Satlas) to take advantage of the global representation learning. While ImageNet is considered a large dataset with 1 million training images, Satlas is around 100$\times$ larger. At the same time, Satlas is curated for remote sensing data, but ImageNet is a general dataset that covers multiple objects (e.g., animals, musical instruments, plants) that are not typically seen from satellite images.

While Swin-T (pre-trained with Satlas) performs well on simultaneously predicting water, cloud, cloud shadow, terrain shadow, and snow/ice, it should be noted that it is also larger and slower than other architectures (Table~\ref{table:runtime-results}, Table~\ref{table:memory-results}). Depending on the application, it would be advantageous to also consider a multi-task MobileNetv3 model which is almost as good as Swin-T in terms of masking accuracy, but only runs for a sixth of the time required for Swin-T, and consumes less RAM and storage. DeepLabv3+ can also be another choice, which can perform more accurately than MobileNetv3, but also consumes less RAM and storage than Swin-T. Our work provides an in-depth analysis of the advantages and disadvantages of the different architectures, enabling other researchers to identify a setting that best suits their requirements.

\subsection{Limitations and future work}
DSWx was used as the training data throughout this work. While it can sufficiently identify water and artifacts such as clouds and shadows, there are inherent limitations in the dataset that could affect our model's performance upon deployment. In particular, snow/ice labels can erroneously occur in areas near the equator and in low elevation areas due to coloration from high sediment concentrations or areas in the water where waves are breaking. Cloud labels from DSWx can also be dilated and thus cover areas that are not necessarily cloud. At the same time, there are instances where DSWx labels fail to detect clouds over water, which are mislabeled as non-water.

Furthermore, we recognize that we are, in essence, making a model of a model by predicting DSWx output. Ideally, we would have a robust dataset of high quality labels generated from the ground, or from digitization of HLS data. In this way we could assess our performance better- the best we can do is to recreate DSWx, so any errors there become our errors. Our use of the LANA dataset was therefore purposeful to assess model performance against manually digitized labels, even though the data volumes for LANA are far smaller. We are encouraged by our performance relative to LANA and use these results as strong evidence for our claims of skill for our multi-task model.

Nonetheless, the framework we introduce in this work is broadly applicable for masking applications beyond learning from DSWx. Manual labels (e.g., cloud labels from the LANA dataset), when available, can also be used to fine tune the model. Once an adequate quantity of manually labeled data is available, the same model we introduced could be used, with similar performance as shown in our experiments in Table~\ref{table:lana-metrics}. This fine tuning, coupled with pre-training on large remote sensing datasets such as Satlas and using a sufficiently large architecture such as DeepLabv3+ or Swin-T could serve as a starting point for future research on satellite masking and reflectance-based estimation.

\section{Conclusion}
\label{sec:conclusion}

Experiments in this paper show that our proposed multi-task models do well to supplement global surface water analysis, illustrating the simultaneous identification of different types of pixels (i.e., water, cloud, cloud shadow, terrain shadow, and snow/ice) in satellite images can result in more accurate masks with a faster runtime. We were able to speedup the runtime by as much as 30$\times$ while using less than half of the standard memory requirement. At the same time, we show that the introduction of our multi-task model in a downstream application results in better performance with an RMSE reduction of 2.64 mg/L for SSC estimation. In particular, the replacement of several modules with a single multi-task model for isolating good quality water pixels results in more accurate SSC predictions. 

While we show multiple options and comparisons across different architectures, we recommend future researchers to start with DeepLabv3+ (pre-trained with ImageNet). Its performance on all masking tasks (water, cloud, cloud shadow, terrain shadow, snow/ice) indicate better performance than all baselines, and better than almost all the other multi-task architectures. At the same time, its runtime and memory consumption are reasonable for its performance. From this starting point, we recommend future researchers to evaluate their needs and scale up or down as necessary (e.g., scale up to Swin-T for better performance or scale down to MobileNetv3 for a smaller and faster model).

The framework presented here is an important step for global surface water analysis, and is part of the SWOT mission wrapper Confluence \cite{tebaldi2021confluence} used for SSC estimation. The proposed pipeline will be used to generate reliable, frequent, sediment flux estimations for every river in the SWORD database \cite{altenau2021surface}, based on global satellite observations. While we show results specific for water pixel identification and SSC estimation, our model can be applied to other areas as well that require distinguishing different types of entities in satellite images such as cloud and cloud shadows. The same proposed pipeline can also be used for other global downstream applications as a faster and resource-efficient alternative.

\section*{Acknowledgments}
We would like to thank the Massachusetts Green High Performance Computing Center (MGHPCC) funded by the Mass. Technology Collaborative for the computational resources that supported this study.

 

\bibliographystyle{IEEEtran}
\bibliography{main}




\vspace{11pt}

\vspace{-33pt}
\begin{IEEEbiography}[{\includegraphics[width=1in,height=1.25in,clip]{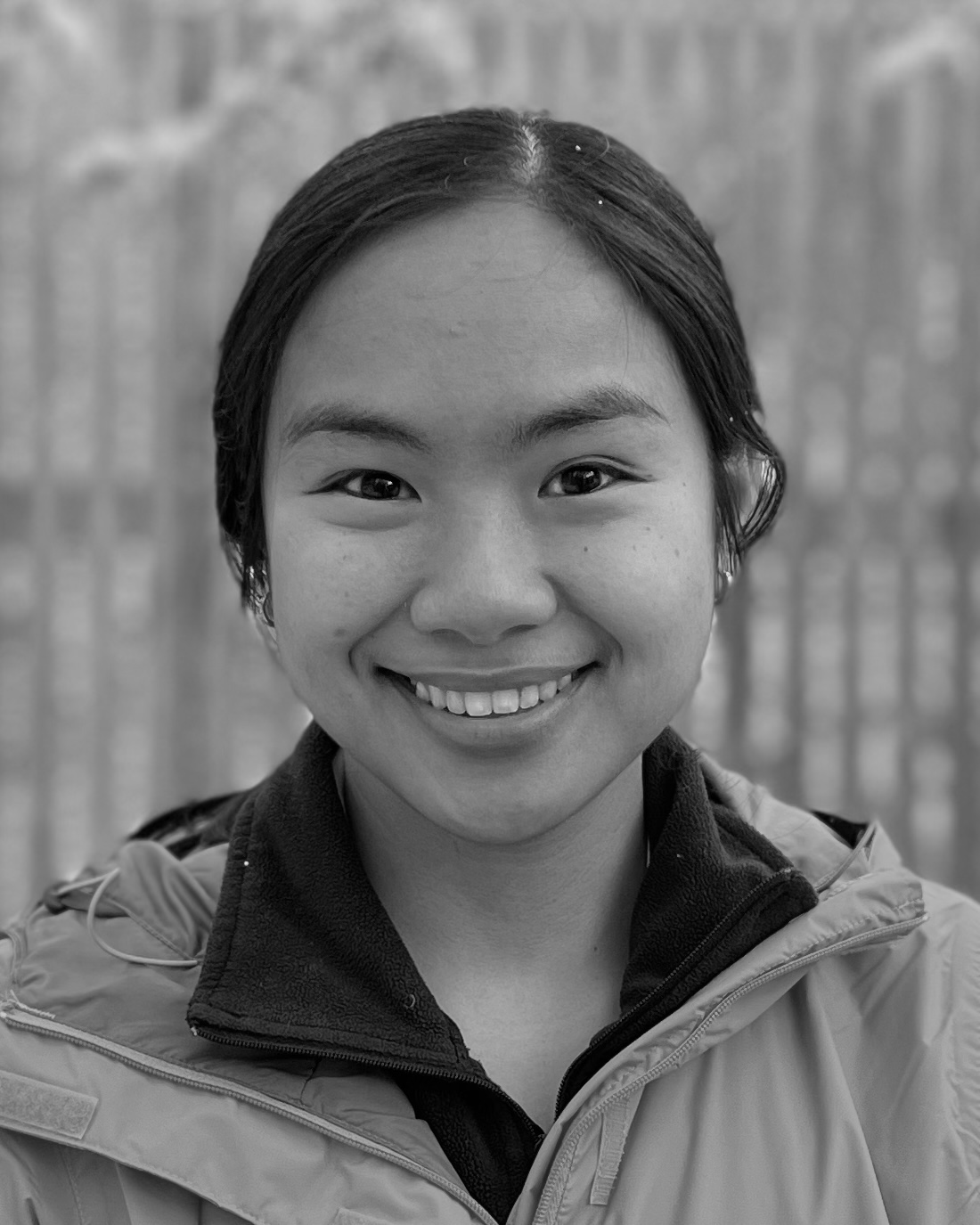}}]{Rangel Daroya} 
received the B.S. degree in electronics and communications engineering and M.S. degree in electrical engineering from the University of the Philippines, Quezon City, Metro Manila, Philippines in 2017 and 2020, respectively.

She is currently working towards a doctoral degree in computer science with the College of Information and Computer Sciences, University of Massachusetts, Amherst, MA, USA. Her research interests include computer vision, explainable AI, and remote sensing.
\end{IEEEbiography}

\begin{IEEEbiography}[{\includegraphics[width=1in,height=1.25in,clip]{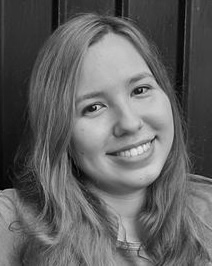}}]{Luisa Vieira Lucchese} 
received the B.S. in civil engineering, and M.S. and Ph.D. degrees in water resources and environmental sanitation from Universidade Federal do Rio Grande do Sul, Porto Alegre, Rio Grande do Sul, Brazil in 2015, 2018, and 2022, respectively.

She is currently a Research Assistant Professor at the Department of Geology and Environmental Science, University of Pittsburgh, Pittsburgh, PA, USA. Her research interests include remote sensing, hydrology, natural hazards, explainable AI, and spatial data science.

\end{IEEEbiography}

\begin{IEEEbiography}[{\includegraphics[width=1in,height=1.25in,clip]{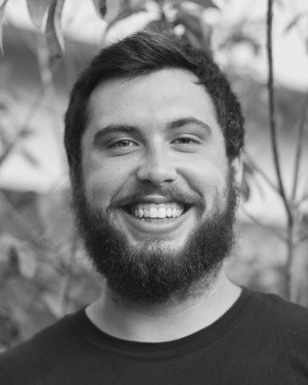}}]{Travis Simmons} 
received the B.S. degree in biology, minors in environmental science and data analytics from College of Coastal Georgia, Brunswick, GA, USA in 2022.

He is currently a Research Fellow at the Department of Civil and Environmental Engineering, University of Massachusetts, Amherst, MA, USA. His research interests include hydrology, data engineering, and pipeline development.
\end{IEEEbiography}

\begin{IEEEbiography}[{\includegraphics[width=1in,height=1.25in,clip]{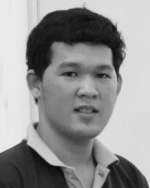}}]{Punwath Prum} 
received the B.S. degree in land management and land administration from the Royal University of Agriculture, Phnom Penh, Cambodia in 2015, and the M.S. degree in environmental observation and informatics from University of Wisconsin, Madison, WI, USA in 2020.

He is currently working towards a doctoral degree with the Department of Geology and Environmental Science, University of Pittsburgh, Pittsburgh, PA, USA. His research interests include studying the impacts of human intervention and climate on water resources using remote sensing, modeling, and field observation.
\end{IEEEbiography}

\begin{IEEEbiography}[{\includegraphics[width=1in,height=1.25in,clip]{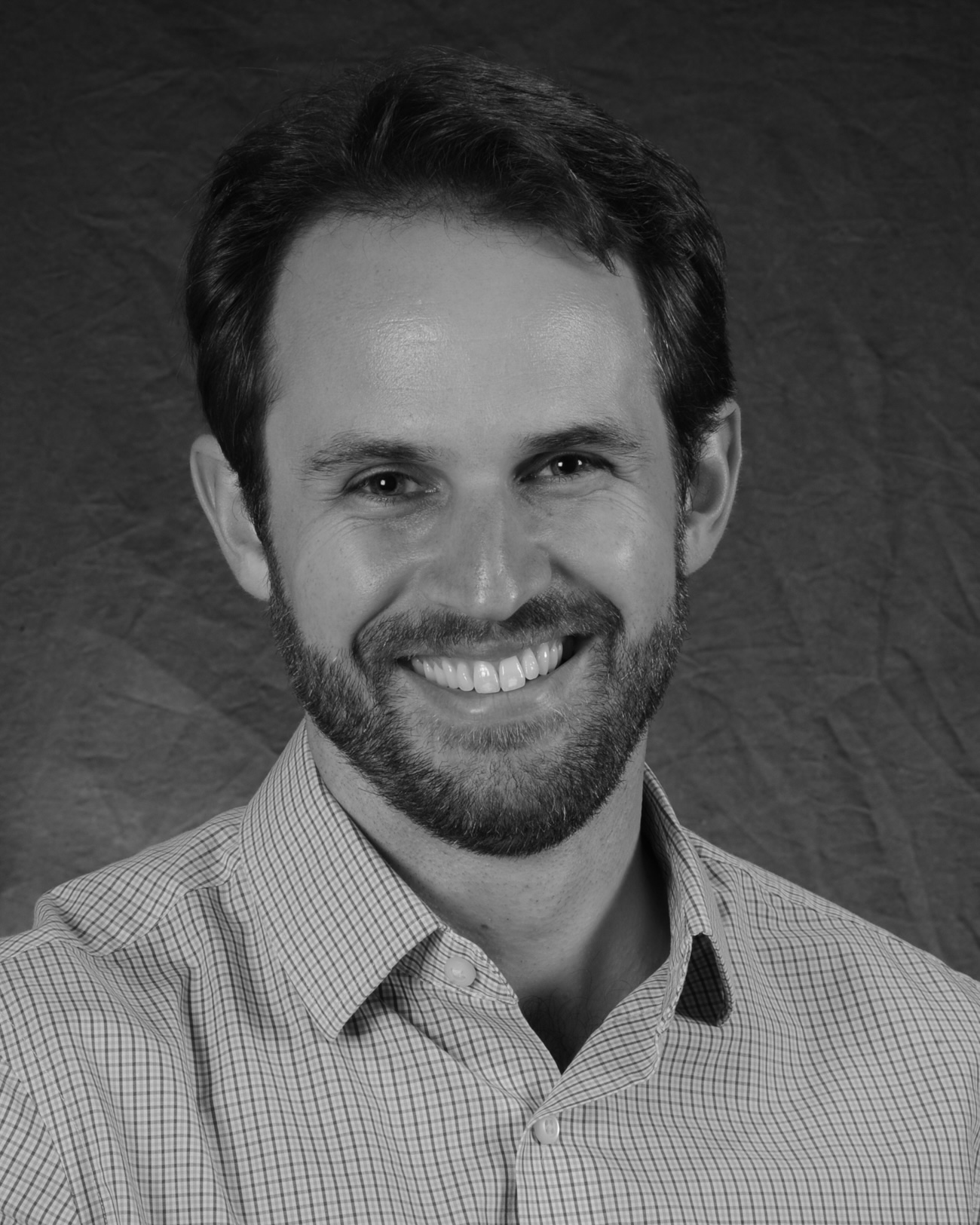}}]{Tamlin Pavelsky} 
received the B.A. degree in geography from the Department of Geography, Middle- bury College, Middlebury, VT, USA, in 2001, and the Ph.D. degree from the University of California, Los Angeles, CA, USA in 2008.

He is currently a Professor at the Department of Earth, Marine and Environmental Sciences, University of North Carolina, Chapel Hill, NC, USA. He is also the U.S. Hydrology Science Lead for the NASA Surface Water and Ocean Topography (SWOT) Satellite Mission. His research interests are focused on the intersections between hydrology, satellite remote sensing, and climate change.
\end{IEEEbiography}

\begin{IEEEbiography}[{\includegraphics[width=1in,height=1.25in,clip]{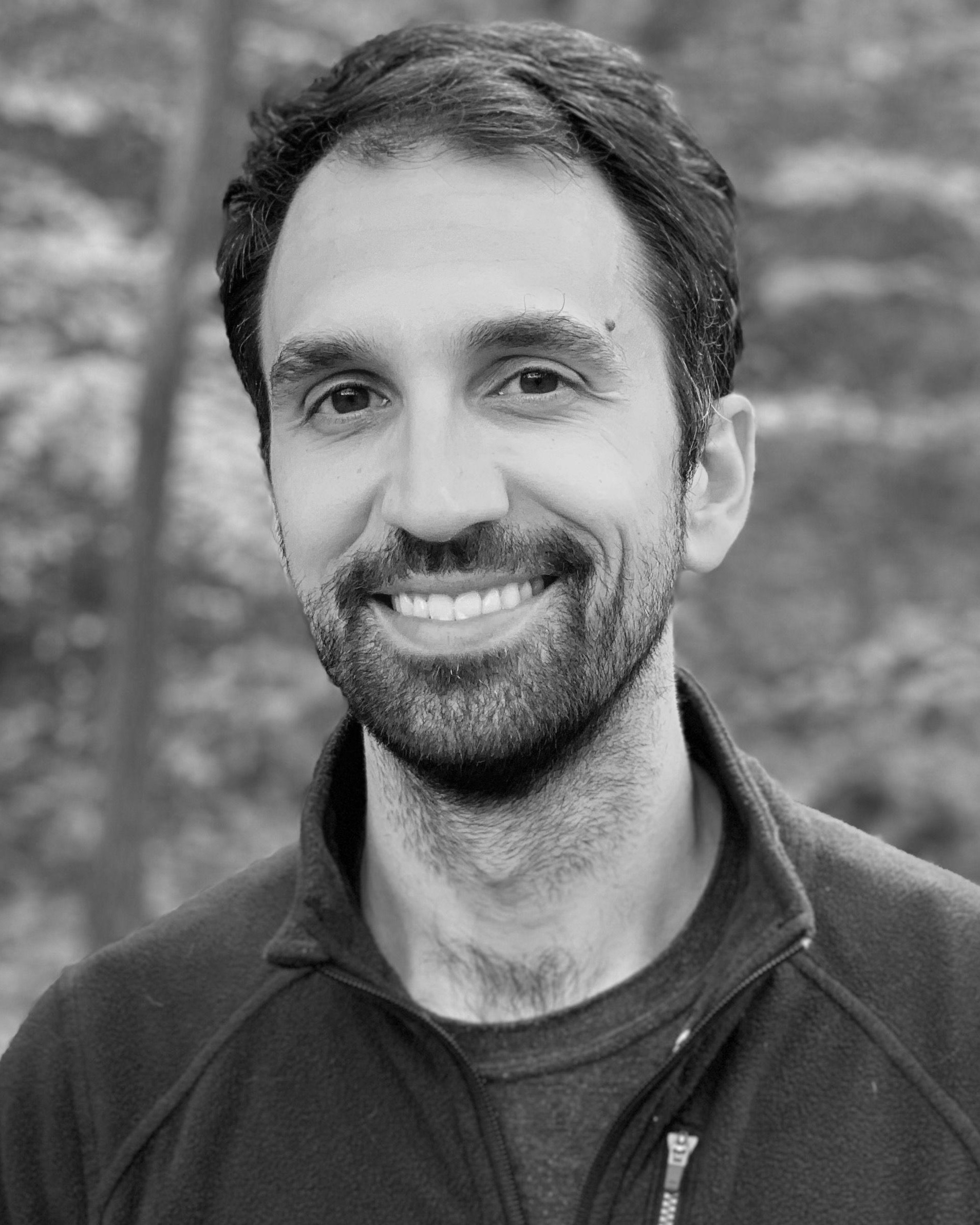}}]{John Gardner} 
received the B.S./B.A. degree in environmental science and geography from the University of Missouri, Columbia, MO, USA in 2010, the M.S. degree from University of Maryland, College Park, MD, USA in 2014, and the Ph.D. degree from Duke University, Durham, NC, USA in 2018.

He is currently an Assistant Professor at the Department of Geology and Environmental Science, University of Pittsburgh, Pittsburgh, PA, USA. He is also the Associate Director at the Pittsburgh Water Collaboratory. His research interests focus on how rivers, lakes, and their landscapes move water, sediment, and elements across continents.

\end{IEEEbiography}

\begin{IEEEbiography}[{\includegraphics[width=1in,height=1.25in,clip]{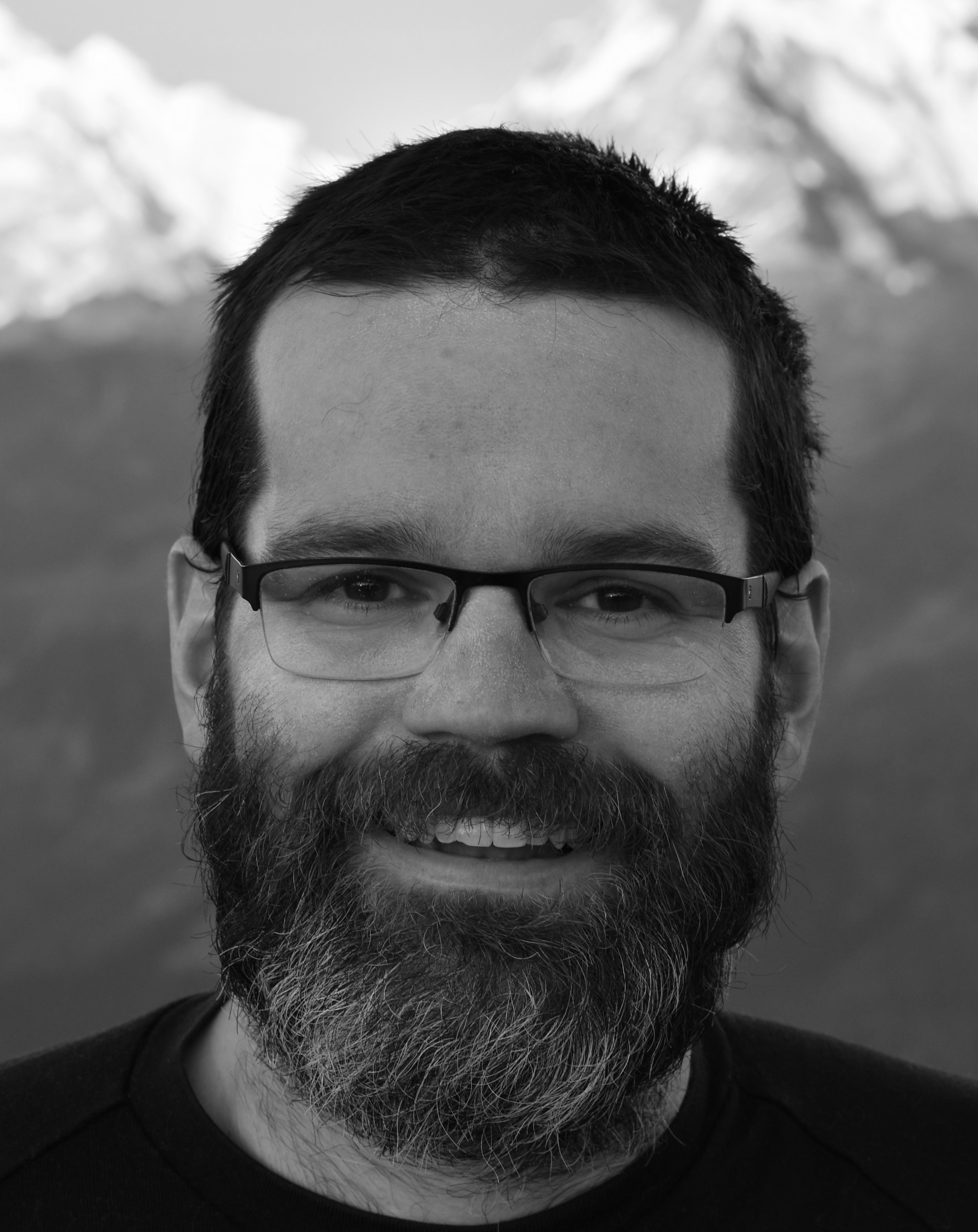}}]{Colin J. Gleason} 
received the B.S. in forest engineering from SUNY Environmental Science and Forestry in Syracuse, New York, USA in 2009. He then received the M.S. in geospatial engineering also at SUNY Environmental Science and Forestry in Syracuse, New York, USA in 2011. He received the Ph.D. degree in geography at UCLA in Los Angeles, CA in 2016.

He is currently the Armstrong Professional Development Professor at the University of Massachusetts, Amherst. Previously, he was an Assistant Professor at the same institution. He is a global hydrologist and geomorphologist that uses primarily satellite data to ask and answer questions about rivers.

Dr. Gleason is a member of American Geophysical Union (AGU) and European Geosciences Union (EGU).
\end{IEEEbiography}

\begin{IEEEbiography}[{\includegraphics[width=1in,height=1.25in,clip]{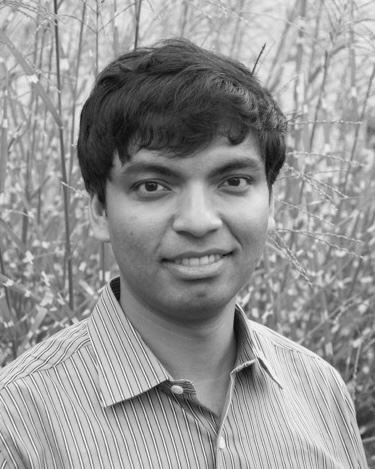}}]{Subhransu Maji} 
received the B.Tech. degree in computer science and engineering from IIT Kanpur, India in 2006, and the Ph.D. degree from University of California, Berkeley, CA, USA in 2011.

He is currently an Associate Professor with the College of Information and Computer Sciences at University of Massachusetts, Amherst. Previously, he was a Research Assistant Professor at Toyota Technological Institute at Chicago. His research interests include computer vision and machine learning. He was the recipient
of a Google graduate fellowship, NSF CAREER Award (2018), and a best paper honorable mention at the IEEE Conference on Computer Vision and Pattern Recognition (CVPR) 2018. He serves on the editorial board of the International Journal of Computer Vision (IJCV).
\end{IEEEbiography}


\vfill

\end{document}